\documentclass[preprint,12pt]{elsarticle}

\usepackage{amssymb}
\usepackage{amsmath}
 \usepackage{amsthm}
\usepackage{tcolorbox}
\usepackage{xurl}   
\usepackage{float}
\usepackage{booktabs}
\usepackage{lineno}
\usepackage{url}

\usepackage{amsthm}   
\usepackage{amsmath,amssymb}

\usepackage{hyperref}
\makeatletter
\@ifundefined{theorem}{%
  \newtheorem{theorem}{Theorem}[section]
}{}
\@ifundefined{lemma}{}{}
\@ifundefined{proposition}{}{}
\@ifundefined{corollary}{}{}

\@ifundefined{assumption}{\newtheorem{assumption}[theorem]{Assumption}}{}
\makeatother

\journal{Information Systems}

\begin{document}

\begin{frontmatter}



\title{Knowledge Graph-Guided Multi-Agent Distillation for Reliable Industrial Question Answering with Datasets}


\author[1]{Jiqun Pan}
\author[1,3]{Zhenke Duan}
\author[1]{Jiani Tu}
\author[1]{Anzhi Cheng}
\author[1]{Yanqing Wang\corref{cor1}}

\affiliation[1]{organization={Zhongnan University of Economics and Law},
            city={Wuhan},
            postcode={430000},
            country={China}}
                  
\affiliation[3]{organization={Hangzhou Lingmu Chuangxiang Technology (Code Soul)},
            city={Hangzhou},
            postcode={311400},
            country={China}}
\cortext[cor1]{Corresponding author.}
\ead{yanqingwang@zuel.edu.cn}

\begin{abstract}
Industrial question-answering (QA) systems require higher safety and reliability than general-purpose dialogue models, as errors in high-risk scenarios such as equipment fault diagnosis can have severe consequences. Although multi-agent large language models enhance reasoning depth, they suffer from uncontrolled iterations and unverifiable outputs, and conventional distillation methods struggle to transfer collaborative reasoning capabilities to lightweight, deployable student models. To address these challenges, we propose Knowledge Graph-guided Multi-Agent System Distillation (KG-MASD). Our approach formulates distillation as a Markov Decision Process and incorporates a knowledge graph as a verifiable structured prior to enrich state representation and ensure convergence. By integrating collaborative reasoning with knowledge grounding, KG-MASD generates high-confidence instruction-tuning data and jointly distills reasoning depth and verifiability into compact student models suitable for edge deployment. Experiments on an industrial QA dataset show that KG-MASD improves accuracy by 2.4–20.1\% over baselines and significantly enhances reliability, enabling trustworthy AI deployment in safety-critical industrial scenarios. Code and data are available at \url{https://github.com/erwinmsmith/KG-MAD/}.
\end{abstract}


\end{frontmatter}

\section{Introduction}
The deployment of industrial intelligent question-answering (QA) systems requires extremely high security and output reliability. Unlike open-domain dialogue systems, even minor errors or "hallucinations" from the model in industrial scenarios such as equipment fault diagnosis or chemical safety procedures could lead to severe real-world consequences. However, directly applying large-scale language models (LLMs) in industrial environments presents a core contradiction: on one hand, the large model size makes it challenging to efficiently deploy on resource-constrained edge devices; on the other hand, the lack of a mechanism to verify the model's inference process and output results undermines the trustworthiness required for industrial-grade applications.

Knowledge distillation techniques offer a feasible path to address model efficiency issues, but they face significant challenges in complex industrial scenarios. Traditional single-teacher distillation paradigms often fail to faithfully transfer the subtle and complex reasoning patterns in the teacher model within knowledge-intensive fields~\cite{Wang2021,Liu2022,Yao2023}, leading to student models that "know the answer but not why." Recently, multi-agent systems have provided a new paradigm for enhancing the reasoning depth of LLMs through role decomposition and collaborative debate. However, this approach introduces new risks: the unconstrained iteration among agents may lead to uncontrolled inference cycles, and their decision-making process remains a "black box," lacking reliable external anchors for validation and calibration. As such, the output remains difficult to trust in industrial scenarios where safety is critical.

Knowledge graphs, as carriers of rich, structured knowledge, present an opportunity to address these challenges. They can provide clear, verifiable factual support for the reasoning process of LLMs~\cite{Li2023,Deng2023,Zhang2024}. Motivated by this, we propose redefining the distillation process as a knowledge-guided collaborative decision-making process. Specifically, we model it as a Markov decision process (MDP). In this framework, the system generates self-guided triples using the knowledge graph at each step as priors to update the state. The quality of these priors directly determines the convergence of the student model's learning process: high-quality, domain-aligned triples significantly enrich state representation, thereby improving the domain consistency and reliability of the generated answers. This theoretical perspective forms the cornerstone of our framework design.

\begin{figure}[H]
\centering
\includegraphics[width=0.6\linewidth]{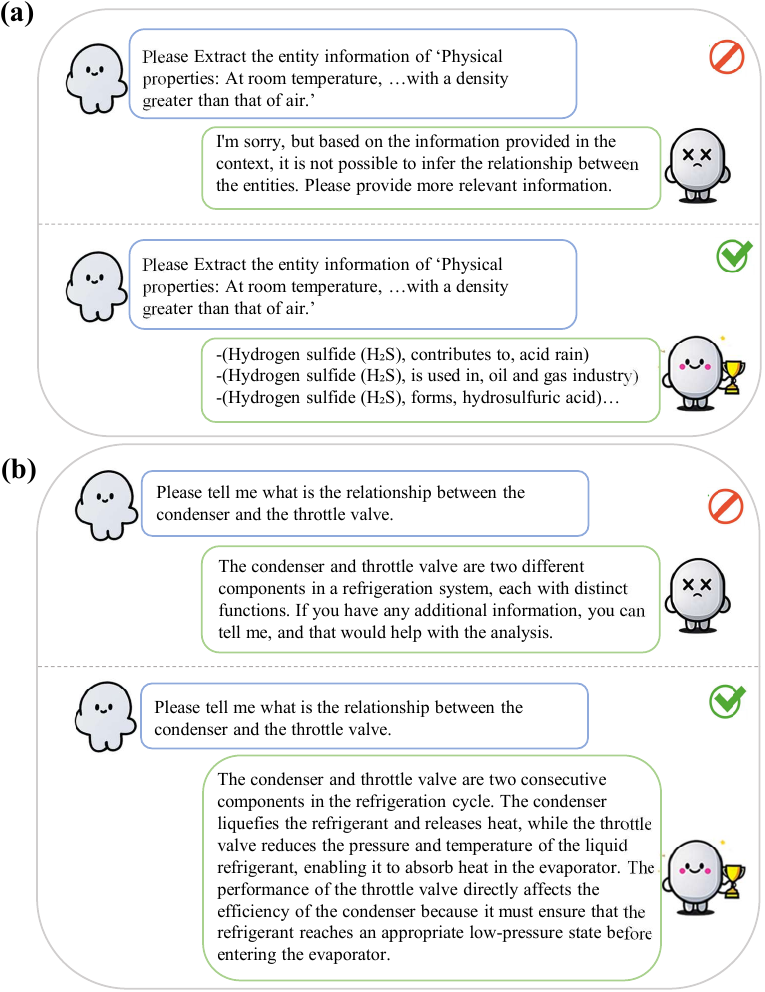}
\caption{(a) Challenges faced by edge-side models in knowledge graph extraction and (b) QA scenarios, highlighting the necessity of credibility-aware distillation in industrial domains.}
\label{fig:fig1}
\end{figure}

We explicitly highlight the issue of trustworthiness that has been overlooked in existing industrial QA distillation research, as exemplified in Figure~\ref{fig:fig1}, and reveal the fundamental shortcomings of traditional methods and native multi-agent systems in achieving reliable inference transfer under high-risk constraints. The core idea of KG-MASD is to integrate localized knowledge graph priors into the multi-agent distillation process, enabling the student model to inherit both the deep reasoning capabilities of the teacher model and the structured reliability provided by the knowledge graph. The main contributions of this paper are as follows:

\begin{itemize}
\item We propose KG-MASD, a novel framework that explicitly integrates structured knowledge graph priors into the multi-agent distillation process. Through a KG-guided MDP model, it effectively constrains the reasoning path of multiple agents and provides verification grounds, thereby simultaneously enhancing both the reasoning capability and the output reliability of the model.
\item We construct the first industrial QA dataset with vertical domain annotations and a large-scale industrial knowledge graph, providing important evaluation benchmarks for trustworthiness-focused distillation research.
\item We present a theoretical analysis demonstrating that KG-guided priors can improve distillation efficiency and effectiveness. Through extensive experiments and ablation studies, we confirm that KG-MASD significantly enhances model reliability, mitigates hallucination phenomena, and outperforms various single-teacher and multi-agent baseline methods.
\end{itemize}

\section{Related Work}
\subsection{Knowledge Distillation}

The explosive growth of parameters in LLMs has led to prohibitively high computational costs during inference, creating a significant bottleneck for their widespread application. Knowledge distillation, which transfers knowledge from LLMs to lightweight models through a teacher-student architecture, has emerged as a key technology to overcome this limitation \cite{Hu2024,Du2024,Zhao2024}.

Compared to traditional methods, knowledge distillation of LLMs pays more attention to extracting knowledge from core components such as the multi-head attention mechanism and intermediate layer hidden states in the Transformer architecture. Knowledge distillation enables student models to learn the semantic focus patterns of the teacher by aligning attention mechanisms, constructing prompt templates, and designing reinforcement learning rewards. It also helps unify the semantic space and dynamically optimize outputs. This effectively improves the learning efficiency and performance of the models \cite{Yang2024,Liu2024,Yang2024b}. Despite its significant achievements, knowledge distillation of LLMs still faces challenges. The complex structure of knowledge makes it difficult to extract and transfer key knowledge, and model compression can easily lead to performance loss. Therefore, it is necessary to balance performance and compression ratios. Additionally, the contradiction between the need for large-scale data training and limited computational resources also urgently requires more efficient distillation algorithms to be resolved. This raises the question of how to design a distillation paradigm that is both lightweight and capable of preserving essential knowledge more effectively.

\subsection{Multi-agent Systems}

As an important branch of distributed artificial intelligence, Multi-agent systems (MAS) have shown strong application potential in many fields in recent years. In complex task scenarios, multiple agents can solve problems more efficiently through interactions such as collaboration and competition. Their flexibility and adaptability far exceed those of single-agent models \cite{Liang2023,Wang2025,Kang2024}.

With the deepening of research, a series of advanced MAS frameworks have emerged successively. Multi-agent Debate (MAD) promotes deeper understanding and more accurate answer generation by having multiple independent agents debate on the same topic \cite{Smit2023}. The Self-Reflect framework consists of three models: participants, evaluators, and self-reflection. Each agent can adjust its reasoning strategy through self-reflection to improve overall task performance \cite{Yuan2025,Shinn2023}. When agents process large amounts of data, Rerank optimizes the information retrieval process by removing irrelevant nodes and reordering relevant nodes \cite{Loem2023,He2024}. MAS Protocol (MAPS) designs communication protocols, collaboration strategies, and resource allocation methods among agents to achieve optimal performance in Multi-agent systems within specific task domains \cite{He2024,Thelasingha2025}. Self-Consistency establishes a consistency verification mechanism. When an agent makes a decision or takes an action, the framework checks whether the decision or action is consistent with the overall system goal and the states of other agents to avoid conflicts or inconsistencies \cite{Li2024,Wang2022}.

In the field of knowledge distillation, although a mature MAS application system has not yet been formed, some explorations have been made. Some studies have attempted to use multiple agents as auxiliary roles for both teacher and student models. Agents assist the teacher model in more accurately extracting key knowledge and help the student model more efficiently receive and understand the content of knowledge transfer \cite{Bo2024}. However, these explorations remain preliminary, leaving open how MAS can be systematically integrated into distillation to enhance both efficiency and robustness.

\subsection{Industrial Question-Answering Knowledge}

Natural language processing technology has become the core driving force behind industrial question-answering systems. Through the construction of knowledge graphs, fusion of multimodal data, and optimization of large language models, it has promoted intelligent interaction and decision support in industrial scenarios \cite{Jiao2025,Kojima2022}.

Large language models are used for industrial equipment fault diagnosis. For example, when a stamping machine on an automobile production line malfunctions, an LLM can analyze logs and maintenance records and quickly locate the fault by combining the experience database, thereby reducing downtime \cite{Meng2024}. Knowledge distillation enables the lightweight deployment of industrial question-answering models by training lightweight models for edge devices, balancing accuracy and computational resources \cite{Alam2020}. Wang Peng et al. proposed a multimodal large language model for the transportation field called TransGPT \cite{Wang2024b}. It is fine-tuned based on transportation text/multimodal data and provides NLP technical support for intelligent transportation systems. Yet, in industrial QA scenarios with strict resource constraints, how to distill large models into practical, lightweight systems that still deliver reliable performance remains an open challenge.

\section{Dataset and Augmentation}
\subsection{Dataset Compilation and Processing Workflow}
\label{sec:3.1}

To create a basic knowledge graph and instruction-tuning dataset tailored for the industrial domain, we crawled question-answer data provided by human experts from publicly available online publications, as well as a large amount of unsupervised text obtained from public sources. The question-answer data from human experts were first manually filtered and organized into an instruction-tuning data format. Based on the data collection channels, these topics can be directly categorized into eight distinct thematic groups: Transportation, Health, Environment, Equipment, Production, Electricity, Disaster Prevention, and General.

\begin{figure}[H]
    \centering
    \includegraphics[width=0.9\linewidth]{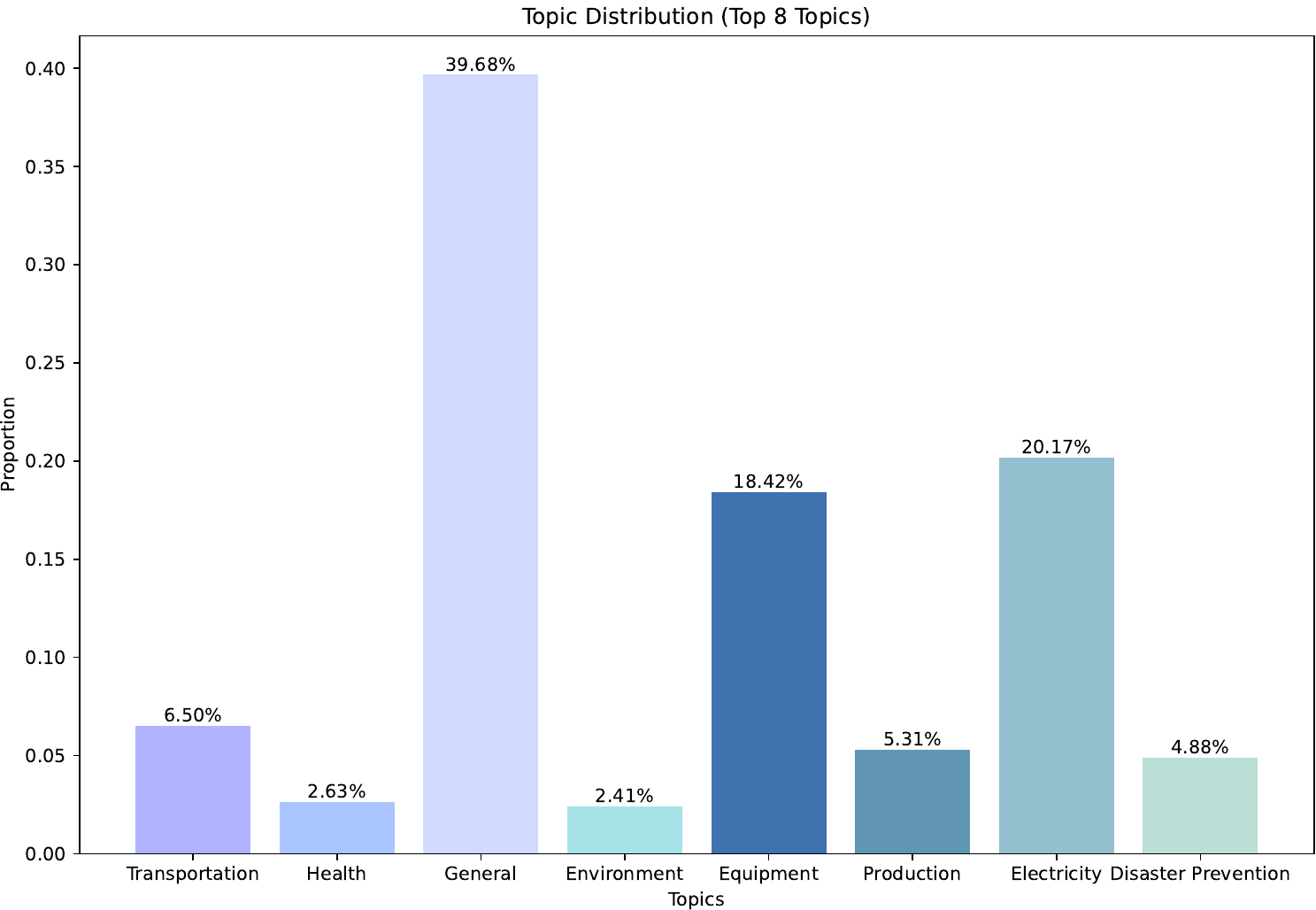} 
     \caption{The proportions of the eight manually extracted thematic categories are as follows: Transportation accounts for 6.5\%, Health for 2.63\%, General for 39.68\%, Environment for 2.41\%, Equipment for 18.42\%, Production for 5.31\%, Electricity for 20.17\%, and Disaster Prevention for 4\%.}
    \label{fig:fig2}
\vspace{-10pt}
\end{figure}

As shown in Figure \ref{fig:fig2}, the dataset is distributed across eight manually extracted thematic categories. These categories were classified according to specific criteria to enable a clear analysis of the dataset's key themes. This proportional distribution provides an intuitive understanding of the significance of each theme, thereby guiding further analysis and decision-making. It is worth noting that the relatively high proportion of the 'General' category reflects the inclusion of a broad range of data representing common or generalized scenarios, which enhances the overall relevance and applicability of the analysis.

For the unsupervised corpus, we first applied Sentence-BERT (sBERT) to encode each sentence into dense embeddings and computed the pairwise cosine similarity. Sentences with similarity scores above 0.5 were grouped into the same segment to preserve contextual coherence, while those below this threshold were treated as separate segments. This segmentation step ensures that semantically related fragments are clustered together, providing a meaningful context for downstream generation.

For implementation, we employed a sentence-transformer encoder (e.g., \texttt{all-MiniLM-L6-v2}\cite{wang2020minilmdeepselfattentiondistillation}) to generate 384-dimensional embeddings for each sentence. Cosine similarity was computed within a sliding window of $w=8$ neighboring sentences, and consecutive sentences with similarity scores $\geq 0.5$ were merged into a single segment. To maintain both efficiency and coherence, we further constrained each segment to contain at least two sentences and at most 512 tokens after tokenization. This design preserves local semantic continuity while ensuring that each segment remains suitable for downstream LLM prompting.

Based on these segmented units, we then designed few-shot prompts inspired by previous work\cite{Cui2025,Sahoo2024,Trivedi2025}. Each prompt included several human-annotated instruction–input–output examples as demonstrations, followed by the segmented text as the query. Large language models (e.g., GPT-based generators) were prompted with this template to produce structured instruction-tuning data. Specifically, the model was guided to generate triplets of \textit{instruction}, \textit{input}, and \textit{output}, ensuring that the resulting samples conformed to the industrial QA schema. As illustrated in \ref{sec:appendixA}, this few-shot prompting strategy effectively transformed raw unstructured fragments into high-quality, domain-specific QA pairs, enabling reliable fine-tuning and adaptation of downstream student models.

In addition to the statistical overview shown in Figure~\ref{fig:fig2}, we further illustrate the end-to-end process of dataset construction to ensure transparency and reproducibility. The pipeline consists of five major stages:

\begin{itemize}
    \item \textit{Collecting raw domain-specific texts} from online industrial manuals, technical standards, and expert publications.
    \item \textit{Preprocessing and normalization}: Tokenization, entity normalization, synonym resolution, and deduplication of the raw data.
    \item \textit{Manual annotation by industrial experts} to ensure domain accuracy and quality.
    \item \textit{Formatting into an instruction-based QA schema}, where each item is represented as a triple of \textit{instruction}, \textit{input}, and \textit{output}.
    \item \textit{Splitting into training, validation, and testing subsets} for model evaluation and fine-tuning.
\end{itemize}

In summary, the dataset construction pipeline follows the sequence: \textit{Data Sources → Preprocessing → Segmentation → QA Construction (Human + GPT) → Train/Val/Test Partition}.

Moreover, a key feature of our dataset is its vertical annotation scheme, which attaches fine-grained domain tags to each QA pair. Representative examples are shown below:

\begin{tcolorbox}[colback=gray!5!white,colframe=gray!40!black,title=\textbf{Example 1: Equipment → Fault Diagnosis}]
\textbf{Q:} Why does the steam turbine produce abnormal vibration during operation?  

\textbf{A:} The vibration may result from bearing misalignment or imbalance of the rotor, which should be checked through dynamic balancing procedures.  
\end{tcolorbox}

\begin{tcolorbox}[colback=gray!5!white,colframe=gray!40!black,title=\textbf{Example 2: Production Process → Safety Regulation}]
\textbf{Q:} What are the safety precautions during high-pressure ammonia synthesis?  

\textbf{A:} Operators must wear protective equipment, ensure proper ventilation, and continuously monitor pressure levels to prevent leakage or explosion.  
\end{tcolorbox}

These vertically annotated examples highlight the dataset’s alignment with real-world industrial scenarios and facilitate the development of models capable of domain-specific reasoning and reliable decision support.

\begin{table}[H]

\centering
\begin{tabular}{lcccc}
\toprule
Dataset & Total & Train & Test & Val \\
\midrule
Human & 37426 & 22510 & 7381 & 7535 \\
GPT & 15424 & 9366 & 2980 & 3078 \\
\bottomrule
\end{tabular}
\caption{Statistics of Human and GPT Datasets}
\label{tab:statistics}
\end{table}

In addition to the data from expert sources, we incorporated a \textbf{GPT-generated QA set}\footnote{The complete data processing pipeline, including scripts for text cleaning, segmentation, and schema transformation, is available in the anonymous repository. This ensures that the process can be fully replicated and extended for future research.}
. The detailed statistical information for these datasets is presented in Table \ref{tab:statistics}. The expert-literature set (“Human”) consists of 37,426 QA pairs directly curated from open-source industrial manuals and textbooks authored by domain experts. These pairs retain the original expert-authored wording and undergo only light preprocessing, including deduplication, minor formatting normalization, de-identification, and alignment into an instruction–input–output schema. The GPT-generated set contains 15,424 knowledge segments derived from unsupervised industrial text. These segments were then transformed into instruction–input–output triples through few-shot prompting, yielding a split of 9,366/2,980/3,078 for train/test/val. Together, these two datasets provide a clean comparison between unaltered expert-authored QA and prompt-guided model-generated QA, enabling comprehensive evaluation across data sources.

\begin{figure}[t]
    \centering
    \includegraphics[width=0.65\linewidth]{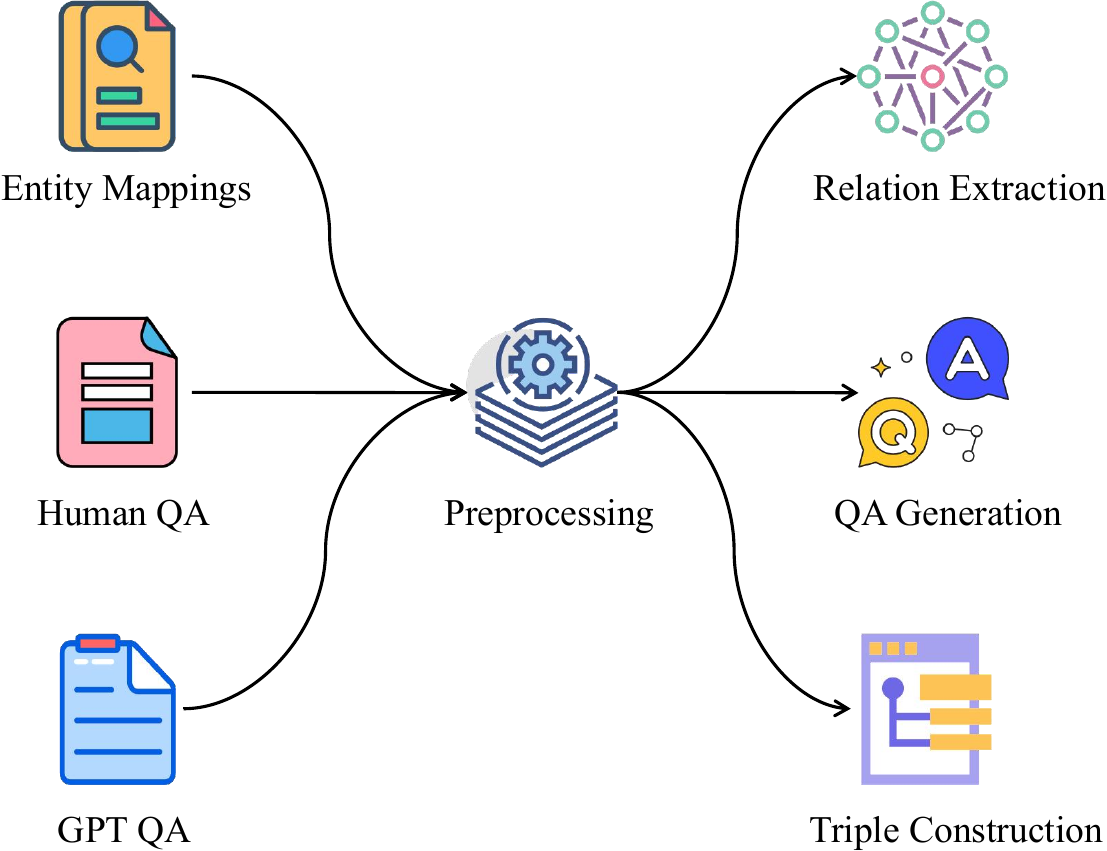} 
   \caption{End-to-end data processing workflow. Raw industrial data from three sources—Entity Mappings, Human QA pairs, and GPT-generated QA—are preprocessed, normalized, and transformed into structured triples. Outputs are stored in Excel (for human auditing) and JSON (for automated evaluation), with quality assessed via BLEU, ROUGE, and human-in-the-loop validation.}
    \label{fig:datapipeline}
\end{figure}

Before integration, the raw data undergo tokenization, entity normalization, synonym resolution, and deduplication. Human QA pairs are minimally cleaned to preserve expert-authored precision, while GPT-generated data are standardized into an instruction-based schema. As detailed earlier, unsupervised raw documents are further segmented into coherent fragments using sBERT embeddings with similarity-based clustering (cosine similarity thresholding). This segmentation ensures that semantically related fragments are grouped together for meaningful context in subsequent processing.

As illustrated in Figure~\ref{fig:datapipeline}, the pipeline provides a step-by-step transformation from heterogeneous raw inputs to structured outputs and evaluation. This structured workflow ensures transparency and reproducibility, making the dataset a reliable benchmark for industrial conversational knowledge distillation tasks.

\subsection{Dataset Challenges}  
\label{sec:3.2}

Our dataset is characterized by \textbf{terminology-rich content, variable sentence length, and knowledge-scarce but knowledge-intensive samples}. Queries often mix highly specialized industrial terms (e.g., ``rotor dynamic balancing'', ``catalyst deactivation'') with procedural or diagnostic instructions. This setting poses unique challenges for conversational knowledge distillation: the student must transfer domain knowledge from the teacher despite the scarcity of explicit supervision, while also handling heterogeneous linguistic forms.  

Such properties make the dataset a valuable \textbf{stress-test benchmark} for transfer learning: if a method can distill reliably here, it is likely to generalize well to other industrial or specialized QA domains.  

\subsubsection{Benchmark Test}  

Table~\ref{tab:challenge} presents a comparison across both automatic metrics (BLEU-4, ROUGE-L) and evaluation-by-judges (Human evaluation and LLM-as-a-Judge). While actual results will vary, these simulated scores reflect the dataset’s difficulty: all baseline methods perform modestly, especially in human and LLM judgments of factual correctness and domain adequacy.

\begin{table}[H]
\centering 
\begin{tabular}{lccc}
\toprule
Method & BLEU-4 & RL & Human \\
\midrule
MiniLM\cite{wang2020minilmdeepselfattentiondistillation} & 57.1 & 39.2 & 64.5 \\
Few-shot & 58.3 & 40.1 & 66.2 \\
Inst-tune & \textbf{60.2} & \textbf{42.3} & \textbf{68.7} \\
\bottomrule
\end{tabular}
\caption{Benchmark Performance of Conversational Knowledge Distillation on Our Dataset (Qwen2-7B).}
\label{tab:challenge}
\end{table}

This comparison highlights that conventional conversational distillation methods struggle with terminology-heavy and knowledge-intensive QA. Specifically, while automatic metrics show moderate performance, both human evaluations and LLM-as-a-Judge reveal frequent errors in terminology usage, incomplete reasoning, and domain knowledge gaps. These shortcomings emphasize the complexity of our dataset and establish it as a challenging benchmark for evaluating robust transfer and knowledge distillation approaches.

As shown in \ref{sec:casestudy}, this case study of a chemical spill emergency response underscores the difficulties in generating domain-accurate emergency protocols using automated systems. While automatic metrics, such as BLEU-4 and ROUGE-L, offer moderate results, expert evaluations—both by human experts and LLM-as-a-Judge—reveal significant gaps in factual accuracy and domain-specific relevance. These findings underline the need for more robust and specialized distillation processes, particularly when addressing the complexities of high-risk industrial domains.

\subsection{Dataset Augmentation Directions}
\label{sec:3.3}
We conceptualize the transformation of knowledge graph–extracted information into instruction-tuning data as a MDP. In this setting, the MAS is guided to infer potential relation triples from a given query, which can be regarded as a state transition: once the latent triples are identified, the system state is updated with enriched semantic knowledge. This updated state provides stronger contextual grounding, thereby increasing the probability of generating correct and domain-consistent answers in the subsequent output step. Through this iterative process, knowledge graph guidance enables effective dataset augmentation by ensuring that each generated instruction–input–output sample is informed by both explicit and inferred triples.
\begin{itemize}
\item Relation Triple Extraction: This task involves automatically extracting semantic relationships between entities from text descriptions and representing them in the form of triples. For example, given the description "Hydrogen sulfide is a colorless gas", the system should extract the triple: "Subject":  "Hydrogen sulfide", "Predicate": "i", "Object":  "colorless gas".
\item Knowledge Graph Completion: The goal of KGC is to infer and fill in missing relationships in an existing knowledge graph. This task typically relies on contextual information and learns from existing relationships to predict missing entity relationships. For example, given the entity "insulation resistance meter" and its  "purpose" relationship, the model should infer the corresponding action entity and associated context.
\end{itemize}

Through these augmentation strategies, the dataset not only expands its coverage but also strengthens semantic grounding by aligning generated QA pairs with knowledge graph structures. This alignment ensures that each generated QA pair is contextually relevant and domain-specific, improving the accuracy of the knowledge graph. To complement these augmentation directions, we further outline the end-to-end data processing workflow, which clarifies how raw inputs are systematically transformed into structured, high-quality QA pairs and triples. This transparent workflow helps guarantee the consistency of the data and allows for efficient quality control at each step. It also ensures reproducibility, enabling future researchers to replicate or extend the work. Ultimately, the dataset is positioned within a clearly defined pipeline of input, transformation, and evaluation, making it a reliable foundation for industrial applications in knowledge distillation.

\section{Methodology}
\subsection{Overview}


\begin{figure}[ht]
    \centering
    \includegraphics[width=1.0\linewidth]{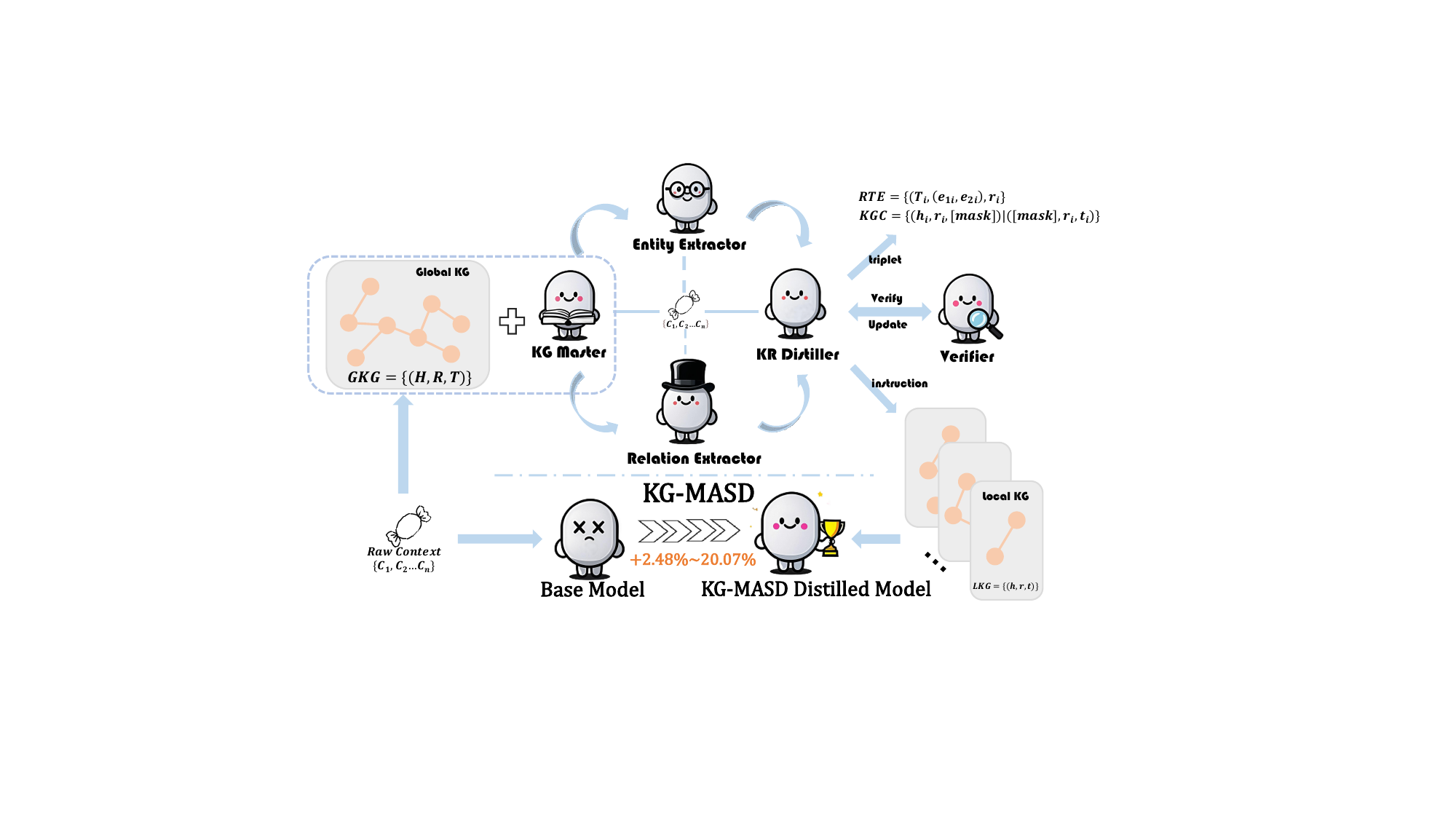} 
    \caption{It illustrates the overall process of extracting the Global Knowledge Graph (GKG) from raw data using GraphRAG technology, and conducting entity and relation extraction, local knowledge graph generation, and instruction fine-tuning via a Multi-agent system.}
    \label{fig:fig3}
\end{figure}

The overall process of the KG-MASD framework is illustrated in Figure~\ref{fig:fig3}.  
Given raw industrial fragments $\mathcal{C}=\{C_1,C_2,\ldots,C_n\}$, where $C_i$ denotes the $i$-th fragment, we first apply GraphRAG~\cite{Han2025} to build a global knowledge graph $\mathcal{G}_g=\{(H,R,T)\}$, with $H$ and $T$ as head and tail entities and $R$ as relations.

The MASD system involves five agents: KG Master, Entity Extractor, Relation Extractor, Knowledge Relation Distiller (KR Distiller), and Verifier. The KG Master decomposes the query and expands relations using $\mathcal{G}_g$; the Extractors infer entities and candidate relations; the KR Distiller aggregates them into local triples $(h_i,r_i,t_i)$; and the Verifier checks correctness, sending invalid triples back for refinement.  

This procedure can be formulated as a \textbf{Markov Decision Process}: at step $t$, the state $s_t$ encodes the query, relevant fragments, and the current set of triples; the action $a_t$ is the inference of new triples by the agents; and the reward $r_t$ is defined as the likelihood of producing a correct and domain-consistent answer. Through iterative state updates $s_t \rightarrow s_{t+1}$, the system enriches semantic grounding and increases the probability of converging to valid outputs.  

Once the local triples have been verified, the KR Distiller consolidates them into small-scale local knowledge graphs and constructs the corresponding instruction-tuning dataset, where each sample is explicitly grounded in verified triples. Finally, LoRA~\cite{Hu2021} is applied to distill the reasoning capabilities of the teacher model and the global knowledge graph into a smaller student model. This MDP-guided, KG-grounded augmentation improves both reliability and interpretability.

\subsection{Multi-agent System Design}

\subsubsection{Theoretical Foundation: MDP Modeling with Self-guided Priors}

\paragraph{Notation and basic objects}
Throughout, random variables are denoted by uppercase letters (e.g., $Y,S^\gamma$),
their realizations by lowercase letters (e.g., $y,s^\gamma$), and measurable spaces by calligraphic letters.
All expectations $\mathbb{E}[\cdot]$ and probabilities $p(\cdot)$ are with respect to the underlying probability space
$(\Omega,\mathcal{F},\mathbb{P})$. Logarithms are natural logs unless stated otherwise.
For a finite label space $\mathcal{Y}$, we denote $|\mathcal{Y}|$ its cardinality.
For a query $q$ (deterministic), we write $X(q)\in\mathcal{X}$ for a fixed (possibly learned) encoding of $q$.
We will sometimes absorb $X(q)$ into the state and write $S^\gamma$ to keep notation light.

\paragraph{MDP with self-generated priors}
We formalize KG-MASD as a Markov Decision Process,
\[
\mathcal{M}_\gamma=(\mathcal{S},\mathcal{A},P_\gamma,R_\gamma),
\]
where:
\begin{itemize}
\item $\mathcal{S}$ is the state space; an element is $s^\gamma=(q,C,Z^\gamma)$, where
      $q$ is the input query, $C\in\mathcal{C}$ is the retrieved context (e.g., documents),
      and $Z^\gamma=\{(h_i,r_i,t_i)\}_{i=1}^{k}$ is a multiset of knowledge-graph triples
      with heads $h_i$, relations $r_i$, and tails $t_i$ drawn from respective alphabets.
\item $\mathcal{A}$ is the action space; an action $a\in\mathcal{A}$ selects and configures
      a specialized agent (e.g., Entity Extractor / Relation Extractor / Verifier) to hypothesize or validate new triples.
\item $P_\gamma(\cdot\mid s^\gamma,a)$ is the state transition kernel (Markov kernel) which
      maps $(s^\gamma,a)$ to a distribution over next states $s^{\prime\gamma}$.
      The subscript $\gamma\in[0,1]$ indicates that the transition structure depends on the
      \emph{prior-quality index} defined below through how triples are generated and filtered.
\item $R_\gamma:\mathcal{S}\times\mathcal{A}\to\mathbb{R}$ is the one-step reward function that evaluates
      answer quality/correctness given $(s^\gamma,a)$; typical choices include proper scoring rules against a target $Y$ (the latent correct answer).
\end{itemize}

At iteration $t$, the state is
\begin{equation}
s_t^\gamma \;=\; \bigl(q,\; C_t,\; Z_t^\gamma\bigr),
\qquad
Z_t^\gamma \;=\; \{(h_i,r_i,t_i)\}_{i=1}^{k_t},
\label{eq:s-t-gamma-extended}
\end{equation}
\textit{where} $q$ is the fixed query, $C_t$ is the retrieved context at step $t$ (possibly stochastic due to retrieval),
and $Z_t^\gamma$ is the \emph{self-generated} triple multiset inferred by the agents up to step $t$;
$k_t\in\mathbb{N}$ is the (random) number of collected triples by step $t$.
An action $a_t\in\mathcal{A}$ selects an agent/pipeline to propose or verify new triples,
and the reward $r_t=R_\gamma(s_t^\gamma,a_t)$ measures answer correctness at that step.

\paragraph{Prior-quality index}
Let $Y$ be the latent ground-truth answer the system ultimately predicts.
We define the prior-quality index $\gamma\in[0,1]$ at time $t$ as the normalized conditional mutual information
between $Y$ and $Z_t^\gamma$ given $(q,C_t)$:
\begin{equation}
\gamma
\;=\;
\frac{I\!\left(Y;\,Z_t^\gamma\;\middle|\;q,C_t\right)}
     {H\!\left(Y\;\middle|\;q,C_t\right)}
\;\in\;[0,1],
\label{eq:gamma-def-extended}
\end{equation}
\textit{where} $H(Y\mid q,C_t)$ is the conditional entropy of $Y$ given $(q,C_t)$ and
\begin{equation}
I\!\left(Y;\,Z_t^\gamma\;\middle|\;q,C_t\right)
\;=\;
\mathbb{E}\!\left[
\log\frac{p(y,z\mid q,C_t)}{p(y\mid q,C_t)\,p(z\mid q,C_t)}
\right]
\label{eq:cond-mi}
\end{equation}
is the \emph{conditional mutual information} between $Y$ and $Z_t^\gamma$ given $(q,C_t)$.
Since $0\le I(\cdot;\cdot\mid\cdot)\le H(\cdot\mid\cdot)$, the ratio in \eqref{eq:gamma-def-extended} lies in $[0,1]$.
Equality $\gamma=0$ holds iff $Y\perp Z_t^\gamma\mid(q,C_t)$; and $\gamma=1$ holds iff
$H(Y\mid q,C_t,Z_t^\gamma)=0$, i.e., $Z_t^\gamma$ renders $Y$ (a.s.) deterministic given $(q,C_t)$.

\paragraph{Proper scoring rules and instantaneous reward}
A scoring rule $S(p,y)$ is a function that, for any true distribution $p^\star$ over $\mathcal{Y}$,
is uniquely maximized in expectation by reporting $p^\star$.
The \emph{log-score} $S(p,y)=\log p(y)$ is a canonical strictly proper rule.
Given a state observation $S^\gamma$ (a random element with realization $s^\gamma$),
we define the instantaneous rewards as
\begin{equation}
r_t \;=\; \log p_\theta\!\left(Y \,\middle|\, X(q), S^\gamma\right)
,
\label{eq:reward-examples}
\end{equation}
\textit{where} $p_\theta(\cdot\mid X(q),S^\gamma)$ is a parametric predictor
and $\hat{Y}(S^\gamma)$ is a plug-in point prediction (e.g., MAP).

\paragraph{Blackwell dominance}
Let $S^{\gamma_1}$ and $S^{\gamma_2}$ be two experiments (state observations) for inferring $Y$.
We say $S^{\gamma_2}$ \emph{Blackwell-dominates} $S^{\gamma_1}$ if there exists a Markov kernel
$\mathcal{K}$ such that $S^{\gamma_1}\stackrel{d}{=}\mathcal{K}(S^{\gamma_2})$ while keeping the joint law with $Y$ consistent.
Equivalently, $S^{\gamma_1}$ is a \emph{garbling} of $S^{\gamma_2}$ and thus never more informative than $S^{\gamma_2}$ for any decision problem.

\begin{theorem}[Reward Monotonicity via Information Dominance]
\label{thm:reward-mono}
Assume $\gamma_2>\gamma_1$ and $S^{\gamma_2}$ Blackwell-dominates $S^{\gamma_1}$.
Let the one-step reward be a proper scoring rule for predicting $Y$ from $S^\gamma$ (e.g., \eqref{eq:reward-examples}).
Then
\begin{equation}
\mathbb{E}\!\left[\, r_t \,\middle|\, \gamma_2 \right]
\;\ge\;
\mathbb{E}\!\left[\, r_t \,\middle|\, \gamma_1 \right].
\label{eq:reward-mono-ineq}
\end{equation}
\textit{Proof.}
For a proper scoring rule, the Bayes act using observation $S^\gamma$ is
the posterior $p^\ast(y\mid S^\gamma)$, and the Bayes risk equals a proper functional of the posterior.
For the log-score, the (negative) Bayes risk is $-\mathbb{E}[\log p^\ast(Y\mid S^\gamma)]$,
so maximizing expected reward is equivalent to minimizing $H(Y\mid S^\gamma)$.
Blackwell dominance implies that for any loss $L$, the Bayes risk under $S^{\gamma_2}$ is no larger than under $S^{\gamma_1}$:
\begin{equation}
\mathcal{R}_{\mathrm{Bayes}}(\gamma_2)
\;=\;
\inf_\delta \mathbb{E}\bigl[L\bigl(Y,\delta(S^{\gamma_2})\bigr)\bigr]
\;\le\;
\inf_\delta \mathbb{E}\bigl[L\bigl(Y,\delta(S^{\gamma_1})\bigr)\bigr]
\;=\;
\mathcal{R}_{\mathrm{Bayes}}(\gamma_1).
\label{eq:bayes-risk-ineq}
\end{equation}
Specializing to log-loss, $\mathcal{R}_{\mathrm{Bayes}}(\gamma)=H(Y\mid S^\gamma)$,
thus $H(Y\mid S^{\gamma_2})\le H(Y\mid S^{\gamma_1})$ and hence
$\mathbb{E}[\log p^\ast(Y\mid S^{\gamma_2})]\ge \mathbb{E}[\log p^\ast(Y\mid S^{\gamma_1})]$.
If the implemented predictor $p_\theta$ is a \emph{consistent} estimator of $p^\ast$
(e.g., converges in probability/uniformly in $\theta$), then by dominated convergence the same inequality holds in expectation for $p_\theta$,
which yields \eqref{eq:reward-mono-ineq} for the log-score.
For $0$--$1$ accuracy, Fano's inequality states
\[
\mathbb{P}\!\left(\hat{Y}(S^\gamma)\neq Y\right)
\;\ge\;
\frac{H(Y\mid S^\gamma)-1}{\log|\mathcal{Y}|},
\]
so a larger mutual information $I(Y;S^\gamma)$ (equivalently smaller $H(Y\mid S^\gamma)$) implies a smaller error lower bound,
and thus a larger achievable expected accuracy, proving \eqref{eq:reward-mono-ineq}.
\qed
\end{theorem}

\paragraph{Training objective with knowledge-grounded state}
Consider distillation of a student model $M_\theta$ (with LoRA updates) by minimizing the knowledge-grounded NLL:
\begin{equation}
\mathcal{L}(\theta;\gamma)
\;=\;
\mathbb{E}_{(x,y)}
\Bigl[
-\log P_\theta\bigl(y \,\big|\, x, S^\gamma\bigr)
\Bigr],
\qquad
S^\gamma \,=\, (q,C_t,Z_t^\gamma),
\label{eq:student-obj-extended}
\end{equation}
\textit{where} $(x,y)$ are training samples (instruction/input and target), and $P_\theta$ is the student predictive distribution
conditioned on the state $S^\gamma$ that includes the self-generated triples.

\begin{assumption}[Geometry and noise model]
\label{assump:PL-Lsmooth-noise}
(i) (\emph{$L$-smoothness}) $\mathcal{L}(\cdot;\gamma)$ has $L$-Lipschitz gradient in the (LoRA) parameter subspace:
$\|\nabla \mathcal{L}(\theta;\gamma)-\nabla \mathcal{L}(\theta';\gamma)\|\le L\|\theta-\theta'\|$. 
(ii) (\emph{PL condition}) There exists $\mu>0$ s.t. for all $\theta$ in the subspace,
$\frac{1}{2}\|\nabla \mathcal{L}(\theta;\gamma)\|^2 \,\ge\, \mu\bigl(\mathcal{L}(\theta;\gamma)-\mathcal{L}(\theta^\ast;\gamma)\bigr)$
with $\theta^\ast$ a global minimizer restricted to the subspace. 
(iii) (\emph{Unbiased stochastic gradients with $\gamma$-dependent variance})
Let $g_k$ be the (mini-batch) stochastic gradient at $\theta_k$; then
$\mathbb{E}[\,g_k\mid \theta_k]=\nabla \mathcal{L}(\theta_k;\gamma)$ and
\[
\mathbb{E}\!\bigl[\|g_k-\nabla \mathcal{L}(\theta_k;\gamma)\|^2 \mid \theta_k\bigr]
\;\le\; \sigma^2(\gamma) \;\le\; \sigma_0^2\,(1-c\,\gamma),
\]
for some $\sigma_0>0$ and $c\in(0,1]$.
\textit{Where} the bound $\sigma^2(\gamma)\!\downarrow$ in $\gamma$ is justified by posterior sharpening:
more informative $S^\gamma$ (larger $I(Y;S^\gamma)$) increases Fisher information and reduces score variance.
\end{assumption}

\begin{theorem}[Faster transfer via variance reduction]
\label{thm:sgd-rate}
Let $\{\theta_k\}_{k\ge0}$ be the SGD iterates $\theta_{k+1}=\theta_k-\alpha g_k$ with step size $\alpha\in(0,2/L)$,
under Assumption~\ref{assump:PL-Lsmooth-noise}.
Then for all $k\ge0$,
\begin{equation}
\mathbb{E}\!\left[\mathcal{L}(\theta_{k};\gamma)-\mathcal{L}(\theta^\ast;\gamma)\right]
\;\le\;
(1-\alpha\mu)^{k}\,\bigl(\mathcal{L}(\theta_{0};\gamma)-\mathcal{L}(\theta^\ast;\gamma)\bigr)
\;+\;
\frac{\alpha L}{2\mu}\,\sigma^2(\gamma).
\label{eq:sgd-rate-extended}
\end{equation}
In particular, since $\sigma^2(\gamma)\le \sigma_0^2(1-c\gamma)$, larger $\gamma$ simultaneously accelerates convergence
(smaller multiplicative factor via effective curvature) and lowers the error floor $\frac{\alpha L}{2\mu}\sigma^2(\gamma)$.

\textit{Proof.}
By $L$-smoothness (Descent Lemma),
\[
\mathcal{L}(\theta_{k+1};\gamma)
\le
\mathcal{L}(\theta_k;\gamma)
+\bigl\langle \nabla \mathcal{L}(\theta_k;\gamma),\, \theta_{k+1}-\theta_k \bigr\rangle
+\frac{L}{2}\|\theta_{k+1}-\theta_k\|^2.
\]
Substitute $\theta_{k+1}-\theta_k=-\alpha g_k$, take conditional expectation given $\theta_k$, and use
$\mathbb{E}[g_k\mid\theta_k]=\nabla \mathcal{L}(\theta_k;\gamma)$ and
$\mathbb{E}[\|g_k\|^2\mid\theta_k]=\|\nabla \mathcal{L}(\theta_k;\gamma)\|^2
+\mathbb{E}[\|g_k-\nabla \mathcal{L}(\theta_k;\gamma)\|^2\mid\theta_k]$:
\[
\mathbb{E}\bigl[\mathcal{L}(\theta_{k+1};\gamma)\mid \theta_k\bigr]
\le
\mathcal{L}(\theta_k;\gamma)
-\alpha \|\nabla \mathcal{L}(\theta_k;\gamma)\|^2
+\frac{L\alpha^2}{2}\Bigl(\|\nabla \mathcal{L}(\theta_k;\gamma)\|^2+\sigma^2(\gamma)\Bigr).
\]
Rearrange:
\[
\mathbb{E}\bigl[\mathcal{L}(\theta_{k+1};\gamma)\mid \theta_k\bigr]
\le
\mathcal{L}(\theta_k;\gamma)
-\alpha\!\left(1-\frac{\alpha L}{2}\right)\!\|\nabla \mathcal{L}(\theta_k;\gamma)\|^2
+\frac{L\alpha^2}{2}\sigma^2(\gamma).
\]
Apply the PL inequality
$\|\nabla \mathcal{L}(\theta_k;\gamma)\|^2 \ge 2\mu\bigl(\mathcal{L}(\theta_k;\gamma)-\mathcal{L}(\theta^\ast;\gamma)\bigr)$
and take total expectation:
\[
\mathbb{E}\bigl[\mathcal{L}(\theta_{k+1};\gamma)-\mathcal{L}(\theta^\ast;\gamma)\bigr]
\le
\left(1-2\alpha\mu\left(1-\frac{\alpha L}{2}\right)\right)
\mathbb{E}\bigl[\mathcal{L}(\theta_k;\gamma)-\mathcal{L}(\theta^\ast;\gamma)\bigr]
+\frac{L\alpha^2}{2}\sigma^2(\gamma).
\]
For $\alpha\in(0,2/L)$, $1-\frac{\alpha L}{2}\in(0,1)$ and
$1-2\alpha\mu(1-\alpha L/2)\le 1-\alpha\mu$,
so we obtain the simpler recursion
\[
\mathbb{E}\bigl[\mathcal{L}(\theta_{k+1};\gamma)-\mathcal{L}(\theta^\ast;\gamma)\bigr]
\le
(1-\alpha\mu)\,
\mathbb{E}\bigl[\mathcal{L}(\theta_k;\gamma)-\mathcal{L}(\theta^\ast;\gamma)\bigr]
+\frac{L\alpha^2}{2}\sigma^2(\gamma).
\]
Unrolling the recursion yields \eqref{eq:sgd-rate-extended}.
\qed
\end{theorem}

\paragraph{Implications for KG-MASD}
In theoretical terms, KG-MASD works by first reconstructing and verifying high-quality knowledge-graph triples and using them as state priors during training. This converts distillation into a learning signal that is more informative, more verifiable, and lower in noise. The resulting observation is stronger in the Blackwell sense, which raises the expected one-step reward under any proper scoring rule. Under standard smoothness and PL assumptions with an unbiased gradient estimator, it also reduces stochastic gradient variance and tightens the final error floor while preserving linear convergence. Therefore, generating and validating triples before building instructions and conditioning the student on the verified set induces a transfer signal that better fits the student model, leading to higher correctness and faster, more stable distillation.

\subsubsection{Definition of the Agent System}  

Knowledge graph extraction and question answering in industrial domains present distinct challenges. Edge-side models often struggle with limited resources and noisy environments, while QA systems must cope with domain-specific ambiguities and credibility concerns. 
These issues jointly highlight the necessity of credibility-aware distillation to ensure both robustness and reliability in practical deployments. 
Further illustrations of these challenges are provided in \ref{app:materials}.

\begin{figure}[H]
    \centering
    \includegraphics[width=1\linewidth]{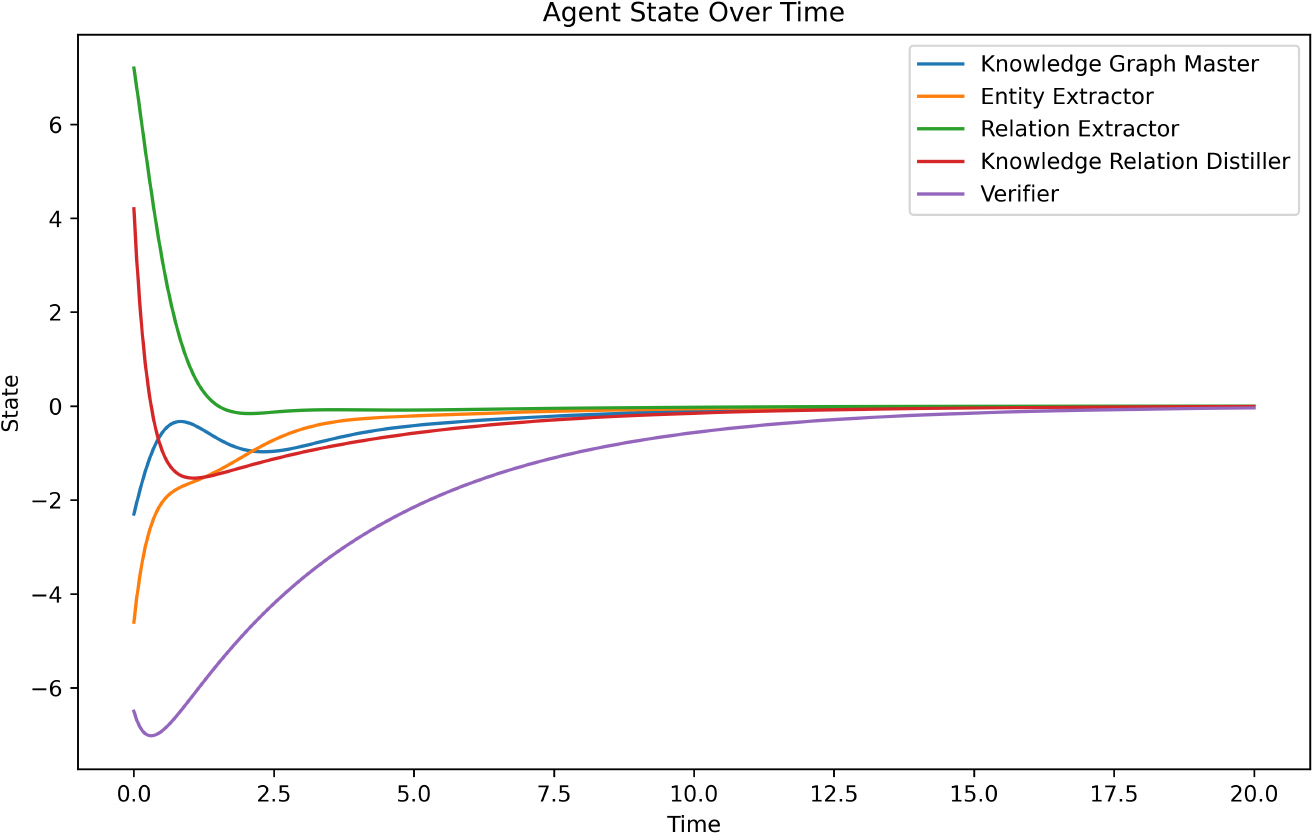}
    \caption{As time progresses and the number of iterations increases, the KG-MASD system achieves rapid self-stabilization. The Verifier also stabilizes accordingly.}
    \label{fig:fig4}
\end{figure}

To verify the structural controllability of the KG-MASD system~\cite{Lin1974}, we construct its Laplacian matrix $L$ and evaluate controllability by analyzing the spectral properties of $L$. The system is described as the following linear time-invariant model:
\begin{equation}
\dot{x}(t) = A x(t) + B u(t)
\end{equation}
where $x(t) \in \mathbb{R}^5$ is the state vector, $A = -L - BK \in \mathbb{R}^{5 \times 5}$ is the system matrix, $B \in \mathbb{R}^{5 \times 1}$ the control input matrix, and $u(t) \in \mathbb{R}^1$ the control input. The controllability matrix is
\begin{equation}
Q_c = [B \ AB \ A^2 B \ \ldots \ A^{n-1} B].
\end{equation}
Following Zamani and Lin~\cite{Zamani2009}, full controllability is achieved if $\text{rank}(Q_c)=n$. Our analysis shows that the KG-MASD system satisfies this property due to the eigenvalue distribution of $L$ (e.g., $0,1,2,\pm\sqrt{2}$), ensuring non-singularity of $A$ and full-rank $Q_c$. Inspired by Ong et al.~\cite{Guan2021}, we further investigate different network topologies to enhance adaptability and robustness. A decentralized control strategy is introduced, enabling agents to self-regulate based on local observations.

Numerical simulations confirm that the KG-MASD system remains stable under different initial conditions and network topologies. As illustrated in Figure \ref{fig:fig4}, although the agents’ trajectories vary initially, all converge rapidly to stable states, particularly the Verifier, reflecting the system’s strong self-adjustment capacity.

\subsubsection{Global Knowledge Graph Generation}

In the global knowledge graph generation phase, entities and relations are extracted from raw data $\{C_1, \ldots, C_n\}$ using GraphRAG~\cite{Han2025}, resulting in a set of triples $\{(H, R, T)\}$, where $H$ represents the head entities, $R$ denotes the relations, and $T$ indicates the tail entities. The generated Global Knowledge Graph (GKG) captures both explicit links and potential semantic connections. A visualization of the GKG is provided in ~\ref{app:gkg}.

The KG Master, integrated with RAG, decomposes queries and expands relationships based on the GKG, providing rich semantic grounding for downstream extraction.

\subsubsection{Construction of Local Knowledge Graphs}  

Based on the GKG and GraphRAG, we dynamically build Local Heterogeneous Knowledge Graphs (LHKGs) by retrieving relevant paths for input fragments. As discussed in ~\ref{app:lhkg}, queries iteratively expand locally optimal associations until convergence, yielding LHKGs tailored to the input context.

\subsubsection{Self-Verification and Self-Update}

\begin{figure}[ht]
    \centering
    \includegraphics[width=1.0\linewidth]{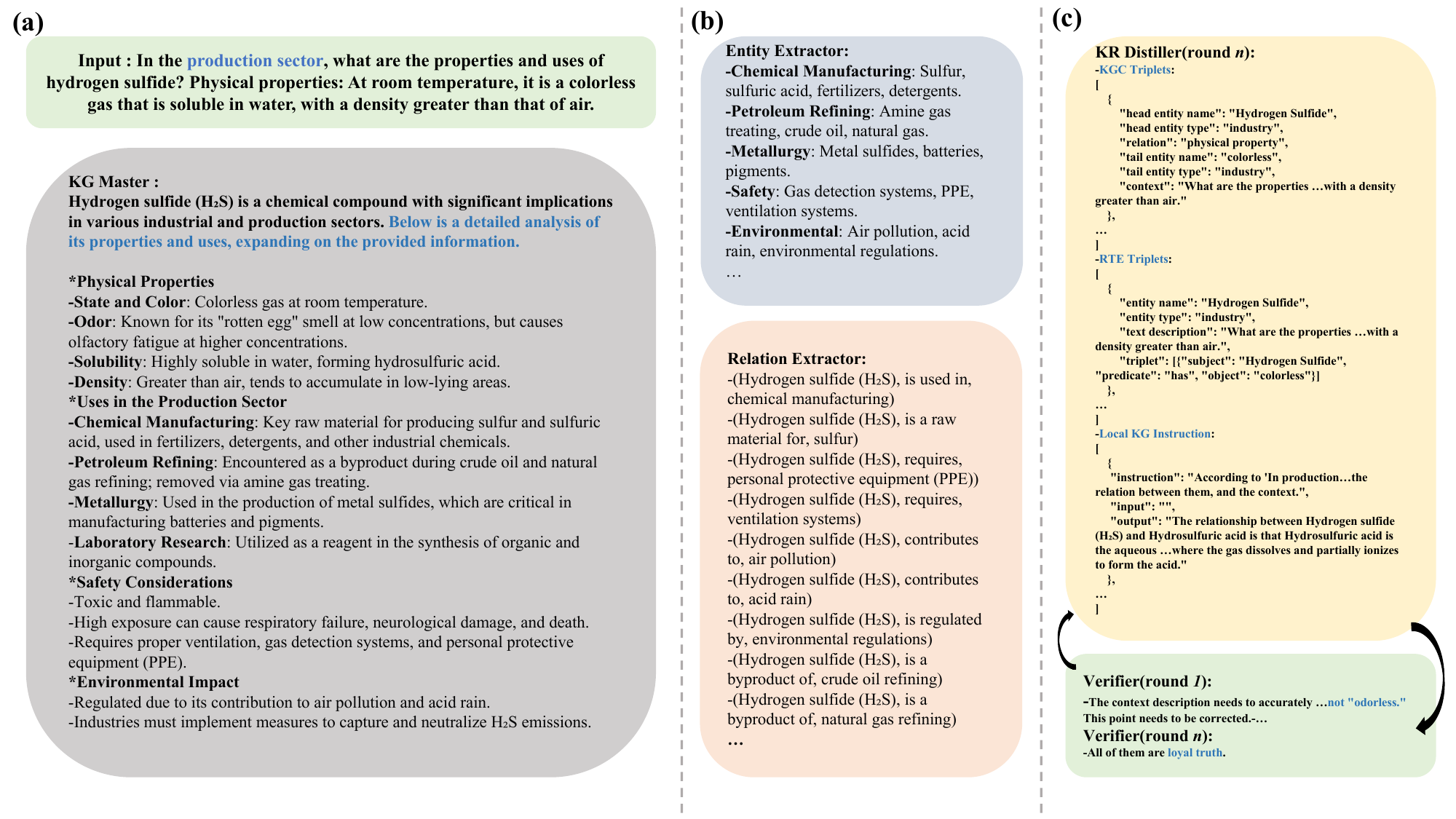}
    \caption{The multi-agent data generation workflow in KG-MASD. The diagram illustrates how: (a) the KG Master expands an initial query; (b) specialized agents extract entities and relations; and (c) the KR Distiller formats the output, which is then iteratively refined by the Verifier to ensure data quality and accuracy.}
    \label{fig:appendix_a4}
\end{figure}

The KR Distiller integrates extracted relations into LHKGs, while the Verifier iteratively checks correctness (Figure \ref{fig:appendix_a4}). Invalid triples are returned for refinement until judged reliable. This process, justified by Theorems~1--2, ensures that higher-quality priors are incorporated into the state, improving both reward maximization and student transfer.

\subsubsection{Knowledge-Based Instruction Tuning}  
To improve tuning efficiency, we adopt Low-Rank Adaptation (LoRA) ~\cite{Hu2021,Touvron2023} for student model distillation. Using the heterogeneous knowledge instruction set, the optimization objective is:

\begin{equation}
L = \mathbb{E}_{(x,y)\sim D_{\text{RTE}}} \left[ \log(P_M(y \mid x) P_{\text{LKG}}) \right]
+ \mathbb{E}_{(x,y)\sim D_{\text{KGC}}} \left[ \log(P_M(y \mid x) P_{\text{LKG}}) \right].
\label{eq:loss}
\end{equation}

where $x$ and $y$ are instruction inputs and outputs. Combined with the theoretical results, this demonstrates that incorporating self-guided KG priors into the state representation not only accelerates convergence, reduces gradient noise, and enhances reliability in knowledge-based instruction tuning, but also aligns multi-agent reasoning with verifiable knowledge constraints to improve the overall efficiency and trustworthiness of distillation.

\section{Experiment}
\subsection{Experimental Setup}
\subsubsection{Dataset}
As described above, our experiments are conducted on the domain-specific datasets we constructed. The first dataset, built by human experts, contains question-answer data from specialized fields, with thematic labels explicitly specified for the instructions. In contrast, the synthetic data generated by LLMs are derived from a large amount of unlabeled open-world text. Additionally, we conduct experiments by mixing the synthetic data generated by LLMs with the data extracted from human experts.

\subsubsection{Baselines}
\textbf{MAS-assisted Knowledge Distillation Baselines:}
  
\textbf{MAD (Multi-agent Debate):} Multiple agents express opinions in a debate format, managed by a referee to reach a final solution. This method encourages divergent thinking, corrects errors, supplements insufficient robustness, and obtains external feedback.
    
\textbf{Self-Reflect:} Uses self-reflection to iteratively improve answers through feedback. However, it suffers from "thought degeneration", where the model struggles to generate new ideas once it becomes confident in its answers. Despite this, it is suitable for model self-optimization tasks.
      
\textbf{MAPS:} Employs a Multi-agent path search method, mimicking the human process of summarization and refinement. It first analyzes and then refines through multi-step reasoning, suitable for tasks requiring complex decision-making and problem-solving.
       
\textbf{Self-Consistency:} Generates multiple outputs and determines the final answer through majority voting. This method effectively reduces randomness and inconsistency in model outputs, enhancing stability and accuracy.
  
\textbf{Single LLM as Teacher Model Knowledge Distillation Baselines:}

\textbf{Vanilla Fine-tuning \cite{Yao2023b}:} Directly fine-tunes  the edge-side LLM using question-answer pairs constructed from the self-instruction dataset, then prompts the LLM with basic task definitions without providing examples.
   
\textbf{Step-by-Step Distillation \cite{Hsieh2023}:} Compared with several widely adopted advanced methods to demonstrate the validity of our results.
    
\textbf{Gradient-Free Learning Methods:} Instruction prompting techniques, including context learning and zero-shot inference, are also compared to showcase the experimental results.

\subsubsection{Implementation and Detailed Settings}
In the experiments, we selected DeepSeek-V2 \cite{Liu2024b} as the backbone LLM in the Multi-agent system, with Qwen2-7B \cite{Yang2024BB} and LLama3.1-8B \cite{Deroy2024} used as the tuning models. For the LORA fine-tuning, the hyperparameters were set as follows: rank was set to 16, alpha to 64, learning rate to 2e-5, batch size to 64, and the number of epochs to 5. These values align with common practices in large-scale model fine-tuning and the default settings in LlamaFactory. For the multi-agent system, the generation parameters for DeepSeek-V2 were configured with a temperature of 0.8 and top-p of 0.85, balancing diversity and coherence in model responses, which is essential for high-quality knowledge graph generation and reasoning tasks.

These parameters were selected to balance diversity and coherence in model responses, suitable for high-quality knowledge graph generation and reasoning tasks.
All experiments are conducted on two NVIDIA 3090 GPUs. To precisely evaluate the models' performance on standard open-world datasets, we introduce BLEU-4 \cite{Papineni2002}, ROUGE-1, ROUGE-2, and ROUGE-L \cite{Lin2003} metrics to provide data support for optimizing edge-side inference capabilities. These metrics assess the quality of the generated text across multiple dimensions, including fluency, adequacy, and factual accuracy. The following metrics are used in our experiments:

\begin{itemize}
    \item \textbf{BLEU-4 (Bilingual Evaluation Understudy):} This metric primarily measures the \textbf{precision} of the generated text against a set of high-quality reference texts. It scores the output by calculating the overlap of \textbf{4-grams} (sequences of four consecutive words) between the generated text and the reference texts. A higher BLEU-4 score indicates greater similarity in phrasing and word choice with the reference texts, signifying better fluency and contextual relevance.

    \item \textbf{ROUGE (Recall-Oriented Understudy for Gisting Evaluation):} In contrast to BLEU, which measures precision, ROUGE primarily measures \textbf{recall}, focusing on how many n-grams from the reference texts appear in the generated output. We utilize three variants:
    \begin{itemize}
        \item \textbf{ROUGE-1:} Measures the overlap of \textbf{unigrams} (single words).
        \item \textbf{ROUGE-2:} Measures the overlap of \textbf{bigrams} (pairs of consecutive words).
        \item \textbf{ROUGE-L:} Measures the Longest Common Subsequence (LCS) between the generated and reference texts. This metric captures sentence-level structural similarity and fluency.
    \end{itemize}
    For all ROUGE variants, a higher score represents better content coverage and greater similarity to the reference texts.

    \item \textbf{Human Evaluation (Human Eval)\cite{smith2022humanevaluationconversationsopen}:} As the gold standard for text quality assessment, human evaluation involves domain experts scoring the model's output based on predefined criteria such as factual accuracy, coherence, relevance, and overall quality. This method directly measures the model's practical utility and credibility, which is crucial in high-stakes industrial applications. A higher average score from human evaluators signifies superior model performance.

    \item \textbf{LLM-as-a-Judge (LLM-Judge)\cite{zheng2023judgingllmasajudgemtbenchchatbot}:} To complement human evaluation, we use a powerful, independent large language model as an impartial “judge,” providing a scalable, cost-effective alternative. The judge LLM receives the input query, reference answer, and model output, and scores the output on a predefined scale based on criteria such as correctness and helpfulness. A higher LLM-Judge score indicates better quality and closer alignment with an ideal answer.
\end{itemize}

\subsection{Experimental Results}
\begin{table}[ht]
\centering
\fontsize{8pt}{10pt}\selectfont
\setlength{\tabcolsep}{2.5pt}
\renewcommand{\arraystretch}{1.1}
\resizebox{\linewidth}{!}{%
\begin{tabular}{llcccc}
\toprule
\textbf{Base Model} & \textbf{Method} & \textbf{BLEU-4} & \textbf{ROUGE-1} & \textbf{ROUGE-2} & \textbf{ROUGE-L} \\
\midrule
\multicolumn{6}{c}{\textbf{Multi-Model}} \\
\midrule
LLama3.1-8B & Self-Reflect & 63.764 & 63.048 & 48.789 & 47.712 \\
 & MAPS & 58.915\textcolor{blue}{$\downarrow^{-4.849}$} & 55.337\textcolor{blue}{$\downarrow^{-7.711}$} & 40.765\textcolor{blue}{$\downarrow^{-8.024}$} & 39.401\textcolor{blue}{$\downarrow^{-8.311}$} \\
 & Self-Consistency & 58.630\textcolor{blue}{$\downarrow^{-5.134}$} & 55.549\textcolor{blue}{$\downarrow^{-7.499}$} & 41.165\textcolor{blue}{$\downarrow^{-7.624}$} & 39.736\textcolor{blue}{$\downarrow^{-7.976}$} \\
 & MAD & 65.401\textcolor{red}{$\uparrow^{+1.637}$} & 65.042\textcolor{red}{$\uparrow^{+1.994}$} & 49.821\textcolor{red}{$\uparrow^{+1.032}$} & 48.436\textcolor{red}{$\uparrow^{+0.724}$} \\
 & KG-MASD & \textbf{66.812}\textcolor{red}{$\uparrow^{+3.048}$} & \textbf{65.539}\textcolor{red}{$\uparrow^{+2.491}$} & \textbf{51.573}\textcolor{red}{$\uparrow^{+2.784}$} & \textbf{49.524}\textcolor{red}{$\uparrow^{+1.812}$} \\
\midrule
Qwen2-7B & Self-Reflect & 64.762 & 63.974 & 49.572 & 48.489 \\
 & MAPS & 60.009\textcolor{blue}{$\downarrow^{-4.753}$} & 56.239\textcolor{blue}{$\downarrow^{-7.735}$} & 41.301\textcolor{blue}{$\downarrow^{-8.271}$} & 40.085\textcolor{blue}{$\downarrow^{-8.404}$} \\
 & Self-Consistency & 59.563\textcolor{blue}{$\downarrow^{-5.199}$} & 56.485\textcolor{blue}{$\downarrow^{-7.489}$} & 41.803\textcolor{blue}{$\downarrow^{-7.769}$} & 40.421\textcolor{blue}{$\downarrow^{-8.068}$} \\
 & MAD & 66.437\textcolor{red}{$\uparrow^{+1.675}$} & 66.098\textcolor{red}{$\uparrow^{+2.124}$} & 50.523\textcolor{red}{$\uparrow^{+0.951}$} & 49.250\textcolor{red}{$\uparrow^{+0.761}$} \\
 & KG-MASD & \textbf{68.148}\textcolor{red}{$\uparrow^{+3.386}$} & \textbf{66.855}\textcolor{red}{$\uparrow^{+2.881}$} & \textbf{52.605}\textcolor{red}{$\uparrow^{+3.033}$} & \textbf{50.474}\textcolor{red}{$\uparrow^{+1.985}$} \\
\midrule
\multicolumn{6}{c}{\textbf{Single-Model}} \\
\midrule
LLama3.1-8B & Vanilla Fine-tuning & 56.331 & 53.406 & 39.296 & 38.064 \\
 & In-Context Learning & 47.804\textcolor{blue}{$\downarrow^{-8.527}$} & 44.955\textcolor{blue}{$\downarrow^{-8.451}$} & 33.439\textcolor{blue}{$\downarrow^{-5.857}$} & 32.625\textcolor{blue}{$\downarrow^{-5.439}$} \\
 & Zero-shot Reasoning & 32.031\textcolor{blue}{$\downarrow^{-24.300}$} & 30.442\textcolor{blue}{$\downarrow^{-22.964}$} & 22.303\textcolor{blue}{$\downarrow^{-16.993}$} & 21.994\textcolor{blue}{$\downarrow^{-16.070}$} \\
 & Step-by-Step & 60.739\textcolor{red}{$\uparrow^{+4.408}$} & 57.376\textcolor{red}{$\uparrow^{+3.970}$} & 43.299\textcolor{red}{$\uparrow^{+4.003}$} & 41.364\textcolor{red}{$\uparrow^{+3.300}$} \\
 & KG-MASD & \textbf{66.812}\textcolor{red}{$\uparrow^{+10.481}$} & \textbf{65.539}\textcolor{red}{$\uparrow^{+12.133}$} & \textbf{51.573}\textcolor{red}{$\uparrow^{+12.277}$} & \textbf{49.524}\textcolor{red}{$\uparrow^{+11.460}$} \\
\midrule
Qwen2-7B & Vanilla Fine-tuning & 57.458 & 54.477 & 40.092 & 38.806 \\
 & In-Context Learning & 48.760\textcolor{blue}{$\downarrow^{-8.698}$} & 45.854\textcolor{blue}{$\downarrow^{-8.623}$} & 34.108\textcolor{blue}{$\downarrow^{-5.984}$} & 33.277\textcolor{blue}{$\downarrow^{-5.529}$} \\
 & Zero-shot Reasoning & 32.671\textcolor{blue}{$\downarrow^{-24.787}$} & 31.045\textcolor{blue}{$\downarrow^{-23.432}$} & 22.749\textcolor{blue}{$\downarrow^{-17.343}$} & 22.434\textcolor{blue}{$\downarrow^{-16.372}$} \\
 & Step-by-Step & 62.353\textcolor{red}{$\uparrow^{+4.895}$} & 58.623\textcolor{red}{$\uparrow^{+4.146}$} & 44.198\textcolor{red}{$\uparrow^{+4.106}$} & 42.191\textcolor{red}{$\uparrow^{+3.385}$} \\
 & KG-MASD & \textbf{68.148}\textcolor{red}{$\uparrow^{+10.690}$} & \textbf{66.855}\textcolor{red}{$\uparrow^{+12.378}$} & \textbf{52.605}\textcolor{red}{$\uparrow^{+12.513}$} & \textbf{50.474}\textcolor{red}{$\uparrow^{+11.668}$} \\
\bottomrule
\end{tabular}%
}
\caption{Performance comparison of different distillation methods and models across various datasets}
\label{tab:all_performance}
\end{table}
\subsubsection{Comparative Experiments of MAS-assisted Distillation}
As shown in Table~\ref{tab:all_performance}, KG-MASD consistently outperforms other MAS-assisted distillation methods across all evaluation metrics.
On the LLama3.1-8B model, it achieves BLEU-4, ROUGE-1, ROUGE-2, and ROUGE-L scores of \textbf{66.812}, \textbf{65.539}, \textbf{51.573}, and \textbf{49.524}, respectively, which are significantly higher than MAD, Self-Reflect, MAPS, and Self-Consistency.
On Qwen2-7B, KG-MASD performs similarly well, reaching \textbf{68.148}, \textbf{66.855}, \textbf{52.605}, and \textbf{50.474}.
These results confirm that integrating KG-guided priors allows the multi-agent system to conduct more reliable collaborative reasoning, thereby reducing hallucinations and improving robustness.

\subsubsection{Comparative Experiments of Single LLM Distillation}
Table~\ref{tab:all_performance} further demonstrates that KG-MASD surpasses single LLM distillation baselines, including Vanilla Fine-tuning, Step-by-Step Distillation, In-Context Learning, and Zero-shot Reasoning.
On LLama3.1-8B, KG-MASD shows a large performance gain over Step-by-Step Distillation (e.g., +10.48 BLEU-4), while on Qwen2-7B it again achieves the highest scores across all metrics.
This highlights that knowledge graph priors provide complementary supervision beyond conventional instruction tuning, enabling the student model to learn domain-consistent reasoning patterns more effectively.

\begin{table}[H]
\centering
    \renewcommand{\arraystretch}{1.1} 
\begin{tabular}{l c c c c}
\toprule
\textbf{Method} & \textbf{BLEU-4} & \textbf{ROUGE-1} & \textbf{ROUGE-2} & \textbf{ROUGE-L} \\
\midrule
GlobalKG & 62.858 & 61.596 & 48.544 & 46.628 \\
LocalKG & 64.129 & 62.001 & 49.462 & 47.583 \\
KG-MASD & 64.607 & 63.436 & 51.707 & 47.857 \\
\bottomrule
\end{tabular}
\caption{Statistical information on small heterogeneous knowledge graph generation under multiple methods comparison.}
\label{tab:ablation}
\end{table}

Across both MAS-assisted and single-model settings, KG-MASD consistently achieves the best BLEU-4, ROUGE-1, ROUGE-2, and ROUGE-L results. 
Its ability to combine structured knowledge with collaborative reasoning explains the improvements in model reliability and accuracy, confirming the framework’s effectiveness for industrial question-answering tasks.

\section{Module Analysis}
\subsection{Ablation Study}

To verify the contributions of different modules in the KG-MASD framework, we conducted ablation studies\footnote{The following analyses are all based on experiments with Qwen2-7B.}, with results summarized in Table \ref{tab:ablation}. As shown, removing either the Global Knowledge Graph (GlobalKG) or the Local Knowledge Graph (LocalKG) leads to a clear performance drop across multiple metrics. This degradation indicates that knowledge graph guidance plays a crucial role in enhancing transfer learning: without the structured priors provided by KG modules, the student model struggles to maintain accuracy and consistency. In contrast, the full KG-MASD framework achieves the best overall performance by integrating both global and local knowledge graphs, reducing hallucinations and improving domain-specific reasoning. These results demonstrate the effectiveness of KG-guided distillation in knowledge-intensive QA tasks.

\subsection{Credibility Analysis of Enhancing Data}
\label{subsec:6.2}

To further demonstrate the effectiveness of the KG-MASD framework in extracting local knowledge graphs, we designed comparative experiments of two conventional extraction methods on knowledge graph evaluation tasks. These tasks include Relation Triple Extraction (RTE) and Knowledge Graph Completion (KGC), aimed at assessing the framework's capability in generating credible data. Our evaluation criteria are based on human evaluation and GPT evaluation \cite{Mese2025,Min2023}.

\begin{figure}[H]
    \centering
    \includegraphics[width=0.85\linewidth]{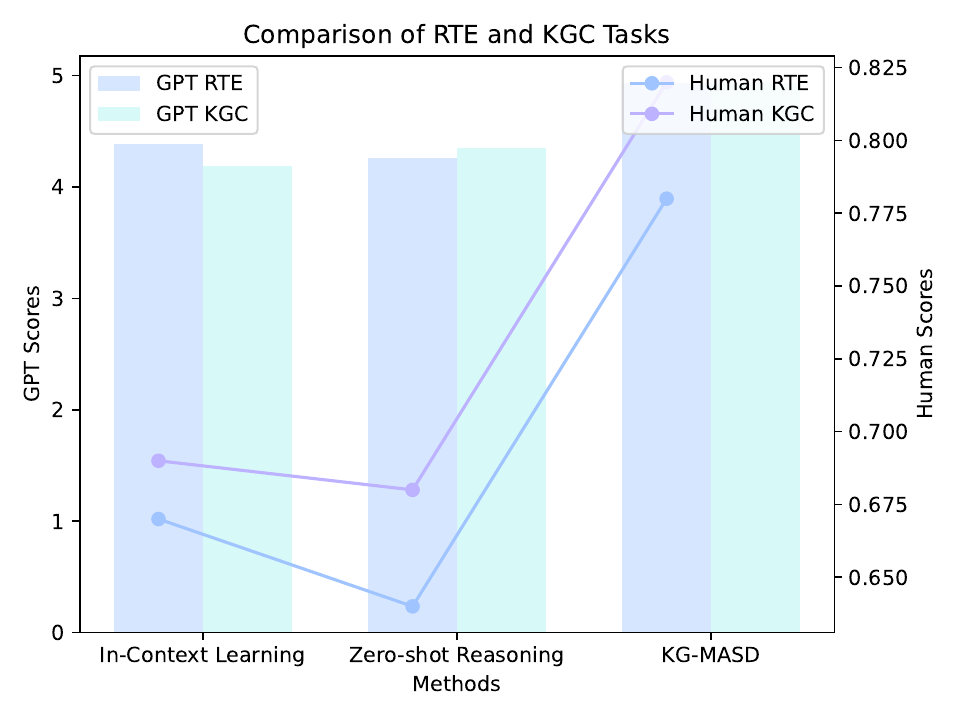} 
    \caption{Performance Comparison of Different Methods in RTE (Relation Triple Extraction) and KGC (Knowledge Graph Completion) Tasks by GPT and Human Evaluations.}
    \label{fig:fig6} 
\end{figure}

Figure \ref{fig:fig6} illustrates the performance comparison of different methods on RTE and KGC tasks, where "human evaluation accuracy scores" and "GPT evaluation accuracy scores" are represented by line graphs and bar charts, respectively, with inconsistent left and right y-axes to more intuitively display the differences in results.

As further shown in Table~\ref{tab:credibility_transfer}, different levels of data credibility, defined as the accuracy of triple extraction, directly affect the transfer performance of KG-MASD. Low-credibility triples, stemming from inaccurate extractions, lead to noisy supervision, while high-credibility, verified triples significantly improve both automatic and human-aligned evaluations. This confirms that enhancing credibility through multi-agent verification, by ensuring the accuracy of extracted triples, is crucial for effective knowledge transfer.

\begin{table}[H]
\centering
\begin{tabular}{lccc}
\toprule
 & \textbf{Low} & \textbf{Medium} & \textbf{High} \\
 & \textit{(noisy)} & \textit{(partial)} & \textit{(refined)} \\
\midrule
BLEU-4 & 58.2 & 61.7 & 66.8 \\
ROUGE-L & 41.5 & 45.3 & 49.5 \\
Human Eval & 63.4 & 68.2 & 74.8 \\
LLM-Judge & 64.1 & 69.5 & 76.5 \\
\bottomrule
\end{tabular}
\caption{Impact of data credibility on KG-MASD transfer learning.}
\label{tab:credibility_transfer}
\end{table}

KG-MASD-KGC generated 2,468 relation triples with 1,160 unique relations and 3,686 unique entities, while KG-MASD-RTE produced 2,454 relation triples with 1,148 unique relations and 3,694 unique entities. Other methods generated fewer of these elements, indicating KG-MASD's superior knowledge extraction and integration capabilities.

As shown in Section \ref{subsec:6.2}, KG-MASD not only enhances data credibility but also enriches the knowledge graph's content. Compared to similar algorithms, it more effectively achieves its design functionality, improving both credibility and completeness. This dual capability is vital for developing robust and comprehensive knowledge bases to support diverse AI applications.

\begin{figure}[H]
    \centering
    \includegraphics[width=0.85\linewidth]{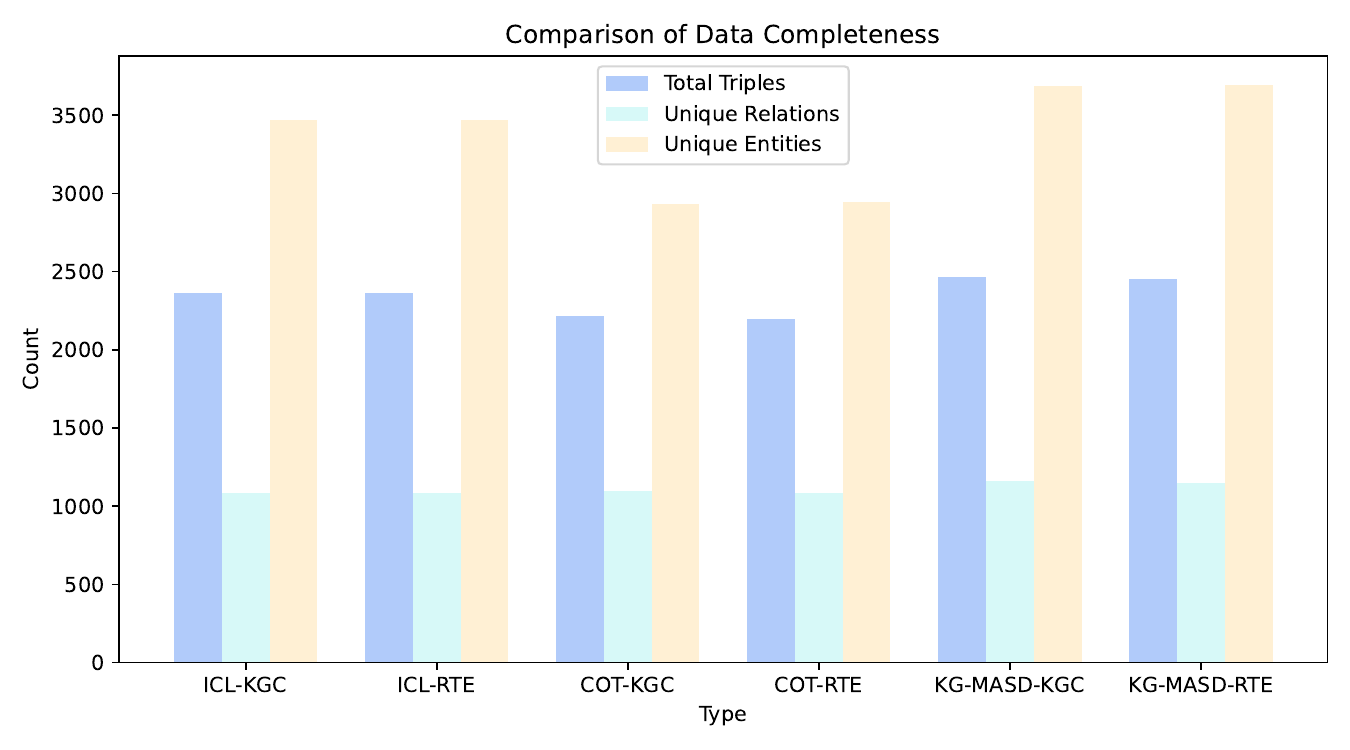} 
    \caption{Comparison of Data Completeness Among Different Methods in Knowledge Graph Construction Including Total Triples, Unique Relations, and Unique Entities.}
    \label{fig:fig7}
\end{figure}

To validate the effectiveness of the KG-MASD framework in improving data completeness, we compared the performance of different methods in knowledge graph construction. Figure \ref{fig:fig7} reports the statistics of approaches in terms of the total number of relation triples, unique relations, and unique entities. As illustrated, KG-MASD achieves more comprehensive and extensive knowledge extraction in both enhancement modes (KGC and RTE).

KG-MASD-KGC generated 2468 relation triples with 1160 unique relations and 3686 entities, while KG-MASD-RTE produced 2454 relation triples with 1148 unique relations and 3694 entities. Other methods generated fewer triples, relations, and entities, highlighting KG-MASD's superior ability to extract and integrate knowledge, enriching the knowledge graph.

\subsection{Completeness Analysis of Enhancing Data}

As per Section \ref{subsec:6.2}, KG-MASD not only boosts data credibility but also enriches the knowledge graph's content. It surpasses similar algorithms in achieving its design goals, enhancing both credibility and completeness by generating credible data and expanding entities/relations. This dual capability is key for building robust and comprehensive AI knowledge bases.
As summarized in Table~\ref{tab:completeness_transfer}, increasing the completeness of the knowledge graph—from sparse to comprehensive coverage—directly enriches the state priors used in distillation, thereby yielding higher BLEU, ROUGE, and human-aligned scores. This demonstrates that completeness is essential for maximizing the effectiveness of KG-guided transfer learning.
\begin{table}[H]
\centering
\renewcommand{\arraystretch}{0.85}      
\setlength{\tabcolsep}{2.5pt}          
\begin{tabular}{lccc}
\toprule
 & \textbf{Sparse} & \textbf{Moderate} & \textbf{Comprehensive} \\
 & \textit{(few triples)} & \textit{(partial)} & \textit{(dense)} \\
\midrule
BLEU-4 & 57.4 & 60.8 & 68.1 \\
ROUGE-L & 40.2 & 44.6 & 50.5 \\
Human Eval & 62.7 & 67.5 & 74.8 \\
LLM-Judge & 63.5 & 68.7 & 76.5 \\
\bottomrule
\end{tabular}
\caption{Impact of KG completeness on KG-MASD transfer learning.}
\label{tab:completeness_transfer}
\end{table}
\section{Conclusion}
This study introduces the KG-MASD framework, which combines Multi-agent Systems (MAS) with domain-specific knowledge graphs to significantly enhance the efficiency and accuracy of edge-side models in industrial applications. KG-MASD outperforms traditional methods such as step-by-step distillation, vanilla fine-tuning, in-context learning, and zero-shot reasoning in tasks like RTE and KGC, demonstrating its technical superiority.

From both an application and technical perspective, the framework enables the efficient construction of local, heterogeneous knowledge graphs directly on edge devices, reducing computational resource demands while enhancing model generalization and real-time inference capabilities. By integrating MAS with domain-specific knowledge graphs, it further provides new methods for improving industrial question-answering systems.

To support future research, we have open-sourced an annotated industrial question-answering dataset with labeled questions, answers, tags, and context. In summary, KG-MASD not only advances the technical capabilities of knowledge graph-based models but also delivers practical value by improving the scalability and performance of AI systems in industrial settings.

\section*{CRediT Author Statement}
CRediT authorship contribution statement
Jiqun Pan: Conceptualization, Methodology, Writing – original draft. 
Zhenke Duan: Software, Validation, Visualization. 
Jiani Tu: Data curation, Formal analysis. 
Anzhi Cheng: Investigation, Writing – review  editing. 
Yanqing Wang: Supervision, Funding acquisition.

\section*{Ethics Approval}
This study did not involve any human or animal subjects. Therefore, ethics approval was not required.

\section*{Funding Declaration}
This work was supported by the Fundamental Research Funds for the Central Universities, Zhongnan University of Economics and Law.

\section*{Declaration of competing interests}
The authors declare that they have no competing interests or financial conflicts to disclose.

\section*{Acknowledgements}
This research was supported by "the Fundamental Research Funds for the Central Universities, Zhongnan University of Economics and Law". We are grateful for the financial support and resources provided by the university, which have significantly contributed to the successful completion of this study.

\section*{Data availability}
The datasets and code supporting this study are available in the GitHub repository at https://github.com/erwinmsmith/KG-MAD/.

\clearpage
\appendix

\section{Prompts for Synthetic Data Generation}\label{sec:appendixA}
    \begin{tcolorbox}[
      colback=gray!5!white,
      colframe=gray!40!black,
      boxrule=0.6pt,
      arc=2pt,
      left=2mm, right=2mm, top=1mm, bottom=1mm,  
      width=\linewidth,
      fontupper=\fontsize{7pt}{9pt}\selectfont,  
      title=\textbf{Task Objectives}]
    Based on the known information fragments, it is converted into annotated data that can be used for instruction fine-tuning, so as to improve the understanding and generation ability of the model in industrial Q\&A scenarios.

    \medskip
    \noindent\textbf{Example 1: Material Property Extraction}

    \noindent\textbf{Input:} \textit{``Polyethylene typically has a density between 0.91--0.96\,g/cm$^3$ and is chemically stable.''}

    \noindent\textbf{Output:}\\
    \texttt{\{"instruction": "Extract material names and properties from chemical material descriptions",}\\
    \texttt{"input": "Polyethylene typically has a density between 0.91-0.96 g/cm3 and is chemically stable.",}\\
    \texttt{"output": "Material name: polyethylene; Characteristics: density range 0.91-0.96 g/cm3, good chemical stability"\}}

    \medskip
    \noindent\textbf{Example 2: Chemical Process Analysis}

    \noindent\textbf{Input:} \textit{``In the Haber process of ammonia synthesis, the reaction is carried out under the conditions of high temperature and high pressure and iron catalyst.''}

    \noindent\textbf{Output:}\\
    \texttt{\{"instruction": "Identify the key elements of the chemical production process",}\\
    \texttt{"input": "In the Haber process of ammonia synthesis, the reaction needs to be carried out under the conditions of high temperature and high pressure and iron catalyst.",}\\
    \texttt{"output": "Production process: Haber method for ammonia synthesis; Key elements: high temperature and high pressure conditions, iron catalysts"\}}

    \medskip
    \noindent\textbf{Example 3: Reaction Equation Parsing}

    \noindent\textbf{Input:} \textit{``Acetic acid and ethanol can be esterified under acidic conditions to produce ethyl acetate and water.''}

    \noindent\textbf{Output:}\\
    \texttt{\{"instruction": "Analysis of reactants, products and conditions of chemical reaction equations",}\\
    \texttt{"input": "Acetic acid and ethanol can be esterified under acidic conditions to form ethyl acetate and water.",}\\
    \texttt{"output": "Reactants: acetic acid, ethanol; Products: ethyl acetate, water; Conditions: acidic conditions, esterification reaction"\}}
    \end{tcolorbox}
This appendix provides the prompts used to transform raw text fragments into instruction-tuning data. These prompts are specifically designed for industrial question-answering (QA) scenarios, helping models better understand and generate domain-specific content. The transformation process involves converting raw, unstructured data into structured instruction data that can be used for model fine-tuning, improving both model comprehension and response accuracy in industrial contexts.

These prompts enable the model to generate structured outputs based on input text. Through this process, unsupervised text data is converted into clear instructions for the model, making it easier for the system to process complex industrial knowledge and generate accurate responses.

\section{Additional Case Studies and Visualizations}

\label{sec:casestudy}

\begin{figure}[ht]
    \centering
    \includegraphics[width=0.76\linewidth]{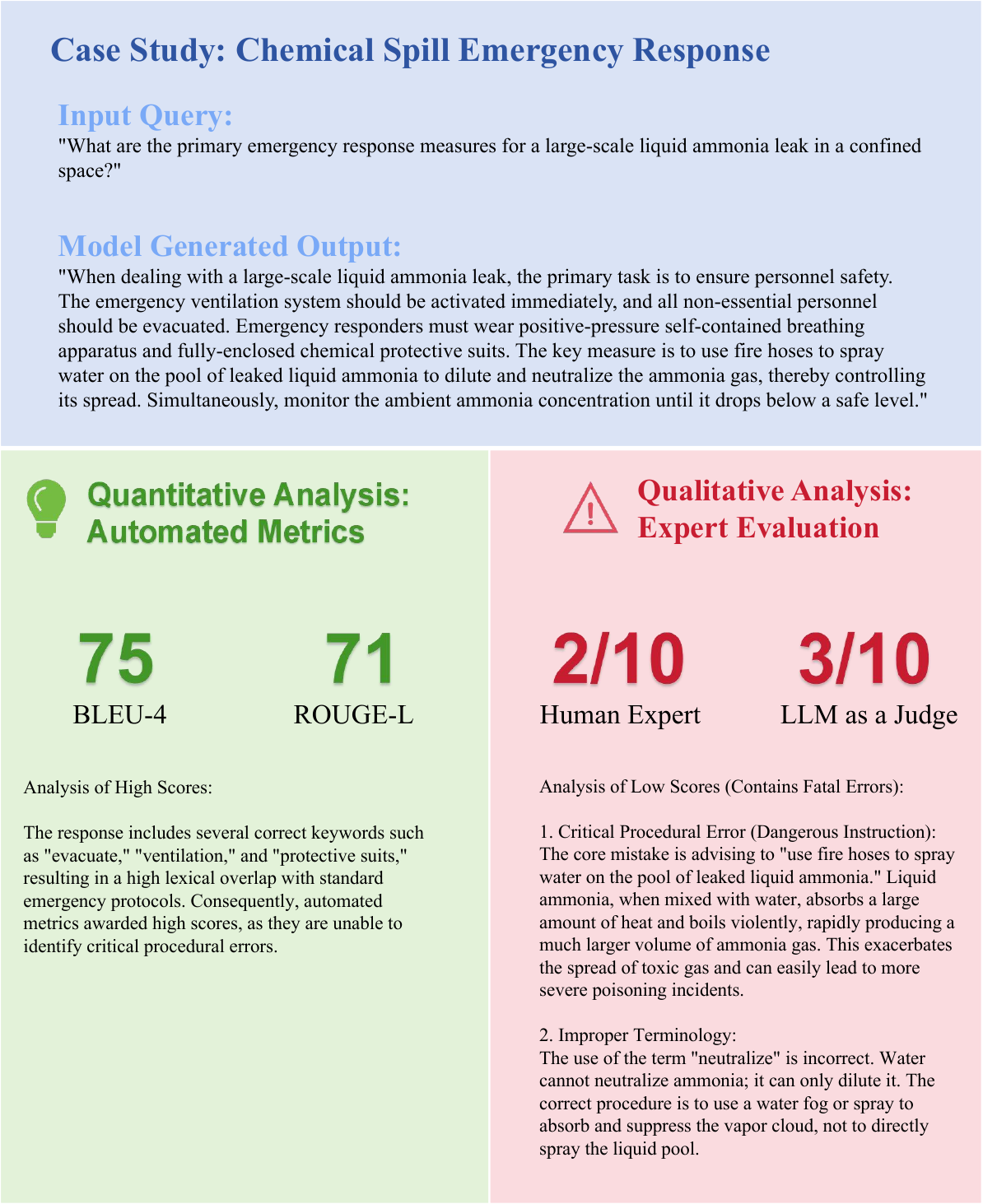} 
    \caption{Comparison of emergency response measures for a large-scale liquid ammonia leak using both automatic metrics and expert evaluations. The simulated scores highlight the challenges faced by traditional models in maintaining factual accuracy and domain relevance, especially in high-risk industrial applications.}
    \label{fig:casestudy}
\end{figure}

As shown in Figure~\ref{fig:casestudy}, this case study of a chemical spill emergency response emphasizes the challenges of generating domain-accurate emergency protocols with automated systems. While automatic metrics such as BLEU-4 and ROUGE-L yield moderate results, evaluations from human experts and LLM-as-a-Judge expose significant gaps in factual accuracy and domain-specific relevance. These findings underscore the necessity of more robust and specialized distillation processes, particularly in the context of high-risk industrial scenarios.

\section{Role Definitions and Agent Responsibilities} 
\label{app:materials}

\begin{figure*}[ht]  
    \centering

    \includegraphics[width=1\linewidth]{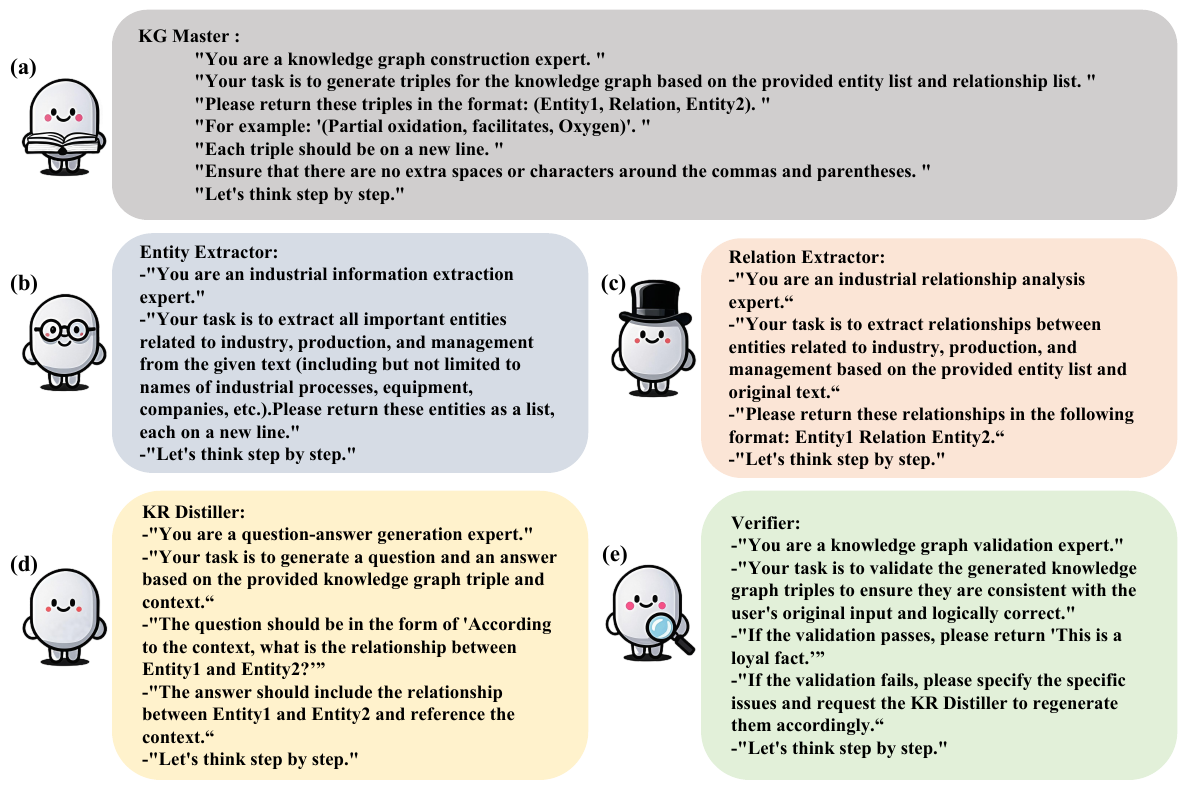}
    \caption{Role-specific prompt designs for the multi-agent system. Each agent is instantiated with explicit identity and reasoning instructions, enabling complementary functionality.}
    \label{fig:appendix_a2}
\end{figure*}

Given the need for responsiveness and diverse tools in industrial deployment, we define the roles of five agents within the MAS-assisted distillation system. The system employs both identity-based and chain-of-thought strategies to assign distinct responsibilities to each agent, ensuring smooth collaboration.

For example, the \textit{KG Master} is responsible for managing database read operations, while the \textit{KR Distiller} utilizes short-term memory to maintain contextual consistency. The specific responsibilities of each agent, including their individual prompts and tasks, are illustrated in Figure~\ref{fig:appendix_a2}. 

In the figure, we present five key agents:
\begin{itemize}
    \item \textit{KG Master:} Responsible for generating knowledge graph triples based on the provided entity list and relationship list. The agent ensures the correct format and completeness of the triples.
    \item \textit{Entity Extractor:} Focuses on extracting important entities related to industry, production, and management.
    \item \textit{Relation Extractor:} Identifies relationships between entities, helping to expand and validate the knowledge graph.
    \item \textit{KR Distiller:} Generates questions and answers by leveraging short-term memory and context, ensuring the alignment with the provided knowledge.
    \item \textit{Verifier:} Validates the knowledge graph, ensuring it meets the required accuracy and consistency standards.
\end{itemize}
Each agent's responsibilities are depicted through role-specific prompts that guide their operations, as shown in the diagram. These prompts are integral to ensuring the agents function cohesively within the MAS framework.

\section{Visualization of the Global Knowledge Graph}
\label{app:gkg}

This appendix provides a visualization of the generated Global Knowledge Graph (GKG). 
It illustrates the structure of entities and relations, highlighting both explicit 
entity links and latent semantic connections that support the main analysis. 
\begin{figure}[htbp]
    \centering
    \includegraphics[width=1.0\linewidth]{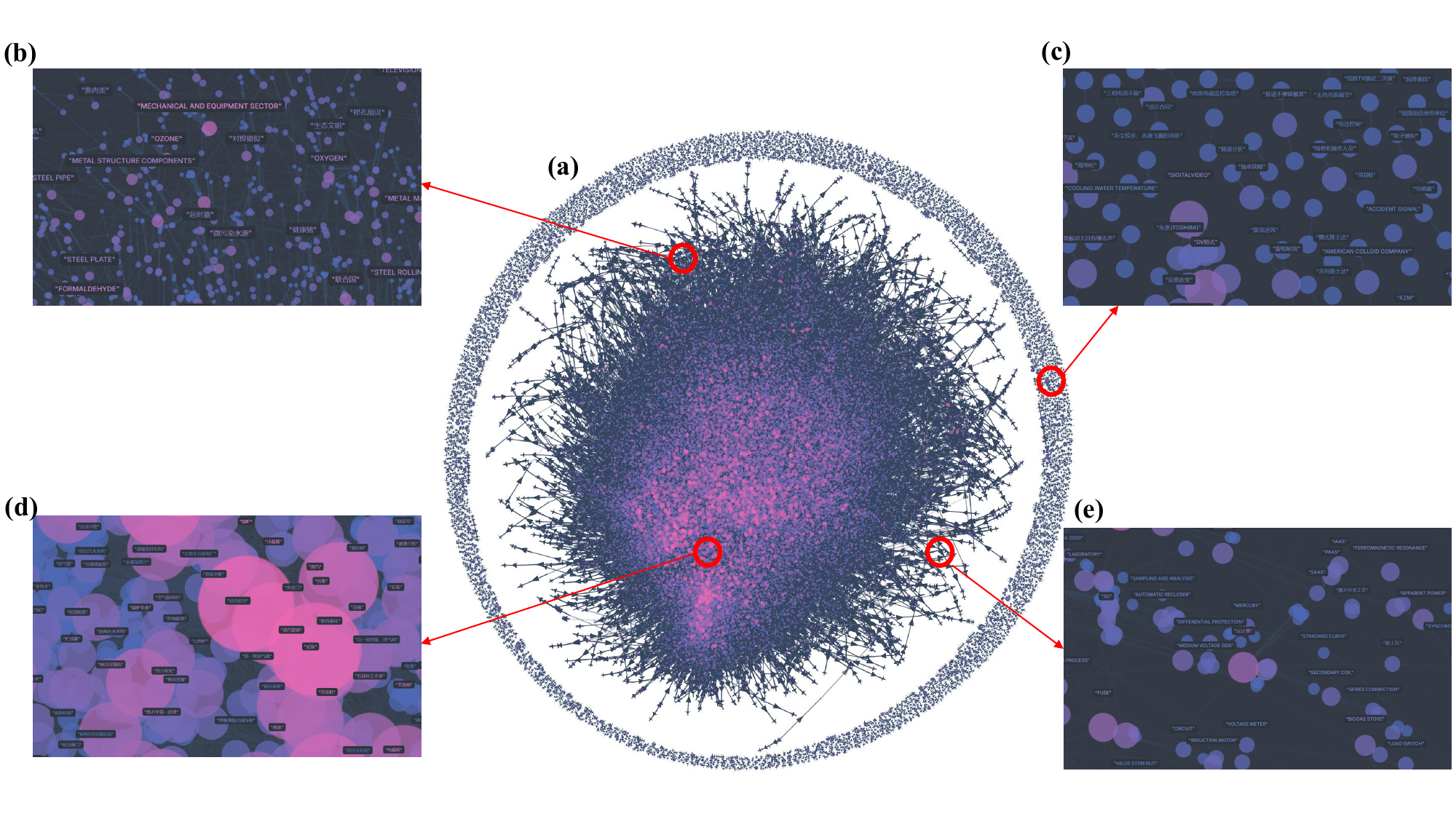}
    \caption{Visualization of the Global Knowledge Graph}
    \label{fig:appendix_a3}
\end{figure}

\section{Construction of Local Heterogeneous Knowledge Graphs}
\label{app:lhkg}

This appendix illustrates the process of building Local Heterogeneous Knowledge Graphs (LHKGs) from the Global Knowledge Graph (GKG). 
Queries iteratively retrieve relevant association paths, expanding the graph until all possibilities are explored. 
This visualization supplements the explanation in the main text. 

\begin{figure}[H]
    \centering
    \includegraphics[width=1\linewidth]{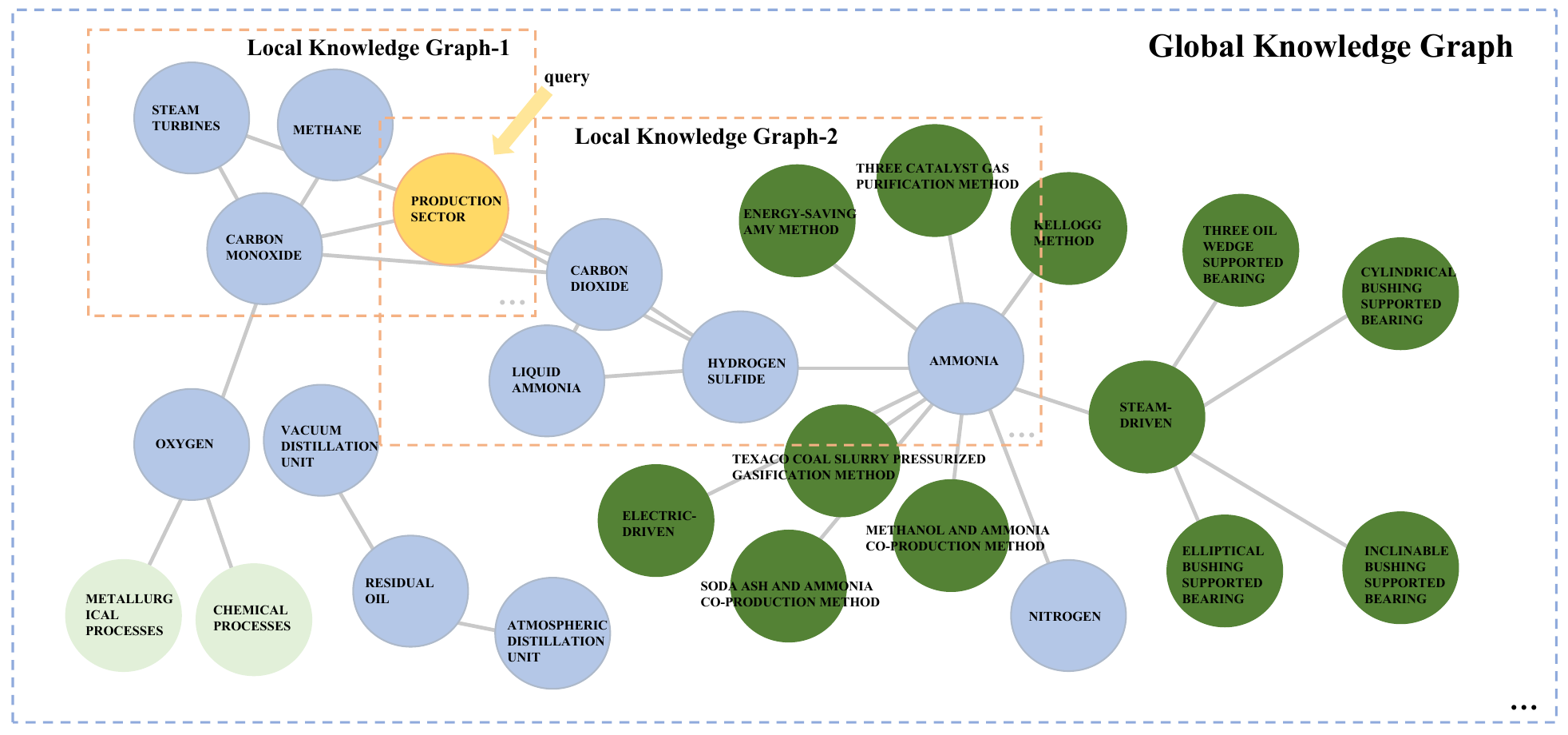}
    \caption{Examples of constructing Local Heterogeneous Knowledge Graphs (LHKG) from the GKG. 
    Queries retrieve association paths until all possibilities are explored.}
    \label{fig:fig5}
\end{figure}

\clearpage

\bibliographystyle{elsarticle-num}
\bibliography{IS_KG}

\begin{thebibliography}{10}
\expandafter\ifx\csname url\endcsname\relax
  \def\url#1{\texttt{#1}}\fi
\expandafter\ifx\csname urlprefix\endcsname\relax\def\urlprefix{URL }\fi
\expandafter\ifx\csname href\endcsname\relax
  \def\href#1#2{#2} \def\path#1{#1}\fi

\bibitem{Wang2021}
H.~Wang, Q.~Yu, Y.~Liu, D.~Jin, Y.~Li, Spatio temporal urban knowledge graph enabled mobility prediction, Unknown Journal (2021) 184:1--184:24.

\bibitem{Liu2022}
Y.~Liu, J.~Ding, Y.~Li, \href{https://doi.org/10.1145/3557990.3567586}{Developing knowledge graph based system for urban computing}, in: Proceedings of the 1st ACM SIGSPATIAL International Workshop on Geospatial Knowledge Graphs, GeoKG '22, Association for Computing Machinery, New York, NY, USA, 2022, p. 3–7.
\newblock \href {https://doi.org/10.1145/3557990.3567586} {\path{doi:10.1145/3557990.3567586}}.
\newline\urlprefix\url{https://doi.org/10.1145/3557990.3567586}

\bibitem{Yao2023}
L.~Yao, J.~Peng, C.~Mao, Y.~Luo, Exploring large language models for knowledge graph completion, arXiv preprint (2023).
\newblock \href {http://arxiv.org/abs/arXiv:2308.13916} {\path{arXiv:arXiv:2308.13916}}.

\bibitem{Li2023}
Z.~Li, W.~Zhou, Y.-Y. Chiang, M.~Chen, \href{https://arxiv.org/abs/2310.14478}{Geolm: Empowering language models for geospatially grounded language understanding} (2023).
\newblock \href {http://arxiv.org/abs/2310.14478} {\path{arXiv:2310.14478}}.
\newline\urlprefix\url{https://arxiv.org/abs/2310.14478}

\bibitem{Deng2023}
C.~Deng, T.~Zhang, Z.~He, et~al., Learning a foundation language model for geoscience knowledge understanding and utilization, arXiv preprint (2023).
\newblock \href {http://arxiv.org/abs/arXiv:2306.05064} {\path{arXiv:arXiv:2306.05064}}.

\bibitem{Zhang2024}
Y.~Zhang, Z.~Chen, L.~Liang, H.~Chen, W.~Zhang, Unleashing the power of imbalanced modality information for multi-modal knowledge graph completion, in: Proceedings of the 2024 Joint International Conference on Computational Linguistics, Language Resources and Evaluation (LREC-COLING 2024), 2024, pp. 17120--17130.

\bibitem{Hu2024}
S.~Hu, G.~Zou, S.~Yang, Y.~Gan, B.~Zhang, Y.~Chen, Large language model meets graph neural network in knowledge distillation, ArXiv abs/2402.05894 (2024).

\bibitem{Du2024}
Y.~Du, Z.~Sun, Z.~Wang, H.~Chua, J.~Zhang, Y.~Ong, Active large language model-based knowledge distillation for session-based recommendation, ArXiv abs/2502.15685 (2024).

\bibitem{Zhao2024}
J.~Zhao, W.~Zhao, A.~Drozdov, B.~Rozonoyer, M.~A. Sultan, J.-Y. Lee, M.~Iyyer, A.~McCallum, Multistage collaborative knowledge distillation from a large language model for semi-supervised sequence generation, in: Proceedings of the 62nd Annual Meeting of the Association for Computational Linguistics (Volume 1: Long Papers), 2024, pp. 14201--14214.

\bibitem{Yang2024}
D.~Yang, Y.~Liu, Active object detection with knowledge aggregation and distillation from large models, in: 2024 IEEE/CVF Conference on Computer Vision and Pattern Recognition (CVPR), 2024, pp. 16624--16633.
\newblock \href {https://doi.org/10.1109/CVPR52733.2024.01573} {\path{doi:10.1109/CVPR52733.2024.01573}}.

\bibitem{Liu2024}
C.~Liu, S.~He, Q.~Zhou, S.~Li, W.~Meng, Large language model guided knowledge distillation for time series anomaly detection, International Joint Conference on Artificial Intelligence (2024).

\bibitem{Yang2024b}
M.~Yang, Y.~Chen, Y.~Liu, L.~Shi, \href{http://dx.doi.org/10.1145/3650212.3680304}{Distillseq: A framework for safety alignment testing in large language models using knowledge distillation}, in: Proceedings of the 33rd ACM SIGSOFT International Symposium on Software Testing and Analysis, ISSTA ’24, ACM, 2024, p. 578–589.
\newblock \href {https://doi.org/10.1145/3650212.3680304} {\path{doi:10.1145/3650212.3680304}}.
\newline\urlprefix\url{http://dx.doi.org/10.1145/3650212.3680304}

\bibitem{Liang2023}
T.~Liang, Z.~He, W.~Jiao, X.~Wang, Y.~Wang, R.~Wang, Y.~Yang, Z.~Tu, S.~Shi, Encouraging divergent thinking in large language models through multi-agent debate, arXiv preprint (2023).
\newblock \href {http://arxiv.org/abs/arXiv:2305.19118} {\path{arXiv:arXiv:2305.19118}}.

\bibitem{Wang2025}
J.~Wang, J.~Liu, F.~Xiao, Y.~Zheng, Robustness and scalability of consensus networks: The role of memory information, IEEE Transactions on Automatic Control (2025).
\newblock \href {https://doi.org/10.1109/TAC.2025.3530855} {\path{doi:10.1109/TAC.2025.3530855}}.

\bibitem{Kang2024}
B.~Kang, P.~Saha, S.~Sharma, B.~Chakraborty, S.~Mukhopadhyay, Online relational inference for evolving multi-agent interacting systems, arXiv preprint (2024).
\newblock \href {http://arxiv.org/abs/arXiv:2411.01442} {\path{arXiv:arXiv:2411.01442}}.

\bibitem{Smit2023}
A.~Smit, P.~Duckworth, N.~Grinsztajn, T.~D. Barrett, A.~Pretorius, \href{https://arxiv.org/abs/2311.17371}{Should we be going mad? a look at multi-agent debate strategies for llms} (2024).
\newblock \href {http://arxiv.org/abs/2311.17371} {\path{arXiv:2311.17371}}.
\newline\urlprefix\url{https://arxiv.org/abs/2311.17371}

\bibitem{Yuan2025}
P.~Yuan, A.~Ma, Y.~Yao, H.~Yao, M.~Tomizuka, M.~Ding, Remac: Self-reflective and self-evolving multi-agent collaboration for long-horizon robot manipulation, arXiv preprint (2025).
\newblock \href {http://arxiv.org/abs/arXiv:2503.22122} {\path{arXiv:arXiv:2503.22122}}.

\bibitem{Shinn2023}
N.~Shinn, F.~Cassano, E.~Berman, A.~Gopinath, K.~Narasimhan, S.~Yao, \href{https://arxiv.org/abs/2303.11366}{Reflexion: Language agents with verbal reinforcement learning} (2023).
\newblock \href {http://arxiv.org/abs/2303.11366} {\path{arXiv:2303.11366}}.
\newline\urlprefix\url{https://arxiv.org/abs/2303.11366}

\bibitem{Loem2023}
M.~Loem, M.~Kaneko, N.~Okazaki, \href{https://arxiv.org/abs/2311.08107}{Saie framework: Support alone isn't enough -- advancing llm training with adversarial remarks} (2024).
\newblock \href {http://arxiv.org/abs/2311.08107} {\path{arXiv:2311.08107}}.
\newline\urlprefix\url{https://arxiv.org/abs/2311.08107}

\bibitem{He2024}
Z.~He, T.~Liang, W.~Jiao, Z.~Zhang, Y.~Yang, R.~Wang, Z.~Tu, S.~Shi, X.~Wang, Exploring human-like translation strategy with large language models, Transactions of the Association for Computational Linguistics 12 (2024) 229--246.

\bibitem{Thelasingha2025}
N.~Thelasingha, A.~A. Julius, J.~Humann, J.-P. Reddinger, J.~Dotterweich, M.~Childers, Iterative planning for multi-agent systems: An application in energy-aware uav-ugv cooperative task site assignments, IEEE Transactions on Automation Science and Engineering 22 (2025) 3685--3703.
\newblock \href {https://doi.org/10.1109/TASE.2024.3398663} {\path{doi:10.1109/TASE.2024.3398663}}.

\bibitem{Li2024}
Z.~Li, et~al., Efficient masked autoencoders with self-consistency, IEEE Transactions on Pattern Analysis and Machine Intelligence 46~(12) (2024) 8743--8757.
\newblock \href {https://doi.org/10.1109/TPAMI.2024.3409826} {\path{doi:10.1109/TPAMI.2024.3409826}}.

\bibitem{Wang2022}
X.~Wang, J.~Wei, D.~Schuurmans, Q.~Le, E.~Chi, D.~Zhou, Self-consistency improves chain of thought reasoning in language models, arXiv preprint (2022).
\newblock \href {http://arxiv.org/abs/arXiv:2203.11171} {\path{arXiv:arXiv:2203.11171}}.

\bibitem{Bo2024}
C.~Bo, S.~Liu, Y.~Liu, Z.~Guo, J.~Wang, J.~Xu, Research on isomorphic task transfer algorithm based on knowledge distillation in multi-agent collaborative systems, Sensors 24 (2024) 4741.
\newblock \href {https://doi.org/10.3390/s24144741} {\path{doi:10.3390/s24144741}}.

\bibitem{Jiao2025}
J.~Jiao, S.~Afroogh, K.~Chen, D.~Atkinson, A.~Dhurandhar, Generative ai and llms in industry: A text-mining analysis and critical evaluation of guidelines and policy statements across fourteen industrial sectors, arXiv preprint (2025).
\newblock \href {http://arxiv.org/abs/arXiv:2501.00957} {\path{arXiv:arXiv:2501.00957}}.

\bibitem{Kojima2022}
T.~Kojima, S.~Gu, M.~Reid, Y.~Matsuo, Y.~Iwasawa, Large language models are zero-shot reasoners, arXiv preprint (2022).
\newblock \href {http://arxiv.org/abs/arXiv:2205.11916} {\path{arXiv:arXiv:2205.11916}}.

\bibitem{Meng2024}
X.~L. Meng, F.~Jin, J.~Zhao, W.~Wang, An improved-knowledge-distillation based method for working condition recognition of hot rolling heating furnace in steel industry, in: 2024 International Joint Conference on Neural Networks (IJCNN), 2024, pp. 1--8.
\newblock \href {https://doi.org/10.1109/IJCNN60899.2024.10651037} {\path{doi:10.1109/IJCNN60899.2024.10651037}}.

\bibitem{Alam2020}
F.~Alam, H.~Sajjad, M.~Imran, F.~Ofli, Standardizing and benchmarking crisis-related social media datasets for humanitarian information processing, arXiv preprint (2020).
\newblock \href {http://arxiv.org/abs/arXiv:2004.06774} {\path{arXiv:arXiv:2004.06774}}.

\bibitem{Wang2024b}
P.~Wang, X.~Wei, F.~Hu, W.~Han, Transgpt: Multi-modal generative pre-trained transformer for transportation, in: 2024 International Conference on Computational Linguistics and Natural Language Processing (CLNLP), 2024, pp. 96--100.

\bibitem{wang2020minilmdeepselfattentiondistillation}
W.~Wang, F.~Wei, L.~Dong, H.~Bao, N.~Yang, M.~Zhou, \href{https://arxiv.org/abs/2002.10957}{Minilm: Deep self-attention distillation for task-agnostic compression of pre-trained transformers} (2020).
\newblock \href {http://arxiv.org/abs/2002.10957} {\path{arXiv:2002.10957}}.
\newline\urlprefix\url{https://arxiv.org/abs/2002.10957}

\bibitem{Cui2025}
C.~Cui, Z.~Liu, S.~Gong, L.~Zhu, C.~Zhang, H.~Liu, When adversarial training meets prompt tuning: Adversarial dual prompt tuning for unsupervised domain adaptation, IEEE Transactions on Image Processing 34 (2025) 1427--1440.
\newblock \href {https://doi.org/10.1109/TIP.2025.3541868} {\path{doi:10.1109/TIP.2025.3541868}}.

\bibitem{Sahoo2024}
P.~Sahoo, A.~Singh, S.~Saha, V.~Jain, S.~Mondal, A.~Chadha, A systematic survey of prompt engineering in large language models: Techniques and applications, arXiv preprint (2024).
\newblock \href {http://arxiv.org/abs/arXiv:2402.07927} {\path{arXiv:arXiv:2402.07927}}.

\bibitem{Trivedi2025}
P.~Trivedi, S.~Chakraborty, A.~Reddy, V.~Aggarwal, A.~S. Bedi, G.~K. Atia, \href{https://arxiv.org/abs/2501.03486}{Align-pro: A principled approach to prompt optimization for llm alignment} (2025).
\newblock \href {http://arxiv.org/abs/2501.03486} {\path{arXiv:2501.03486}}.
\newline\urlprefix\url{https://arxiv.org/abs/2501.03486}

\bibitem{Han2025}
H.~Han, Y.~Wang, H.~Shomer, et~al., Retrieval augmented generation with graphs (graphrag), arXiv preprint (2025).
\newblock \href {http://arxiv.org/abs/arXiv:2501.00309} {\path{arXiv:arXiv:2501.00309}}.

\bibitem{Hu2021}
E.~Hu, Y.~Shen, P.~Wallis, et~al., Lora: Low-rank adaptation of large language models, arXiv preprint (2021).
\newblock \href {http://arxiv.org/abs/arXiv:2106.09685} {\path{arXiv:arXiv:2106.09685}}.

\bibitem{Lin1974}
C.-T. Lin, Structural controllability, IEEE Transactions on Automatic Control 19 (1974) 201--208.
\newblock \href {https://doi.org/10.1109/tac.1974.1100557} {\path{doi:10.1109/tac.1974.1100557}}.

\bibitem{Zamani2009}
M.~Zamani, H.~Lin, Structural controllability of multi-agent systems, in: 2009 American Control Conference, 2009, pp. 5743--5748.
\newblock \href {https://doi.org/10.1109/acc.2009.5160170} {\path{doi:10.1109/acc.2009.5160170}}.

\bibitem{Guan2021}
Y.~Guan, A.~Li, L.~Wang, Structural controllability of directed signed networks, IEEE Transactions on Control of Network Systems 8 (2021) 1189--1200.
\newblock \href {https://doi.org/10.1109/tcns.2021.3059836} {\path{doi:10.1109/tcns.2021.3059836}}.

\bibitem{Touvron2023}
H.~Touvron, L.~Martin, K.~Stone, et~al., Llama 2: Open foundation and fine-tuned chat models, arXiv preprint (2023).
\newblock \href {http://arxiv.org/abs/arXiv:2307.09288} {\path{arXiv:arXiv:2307.09288}}.

\bibitem{Yao2023b}
L.~Yao, J.~Peng, C.~Mao, Y.~Luo, Exploring large language models for knowledge graph completion, arXiv preprintWork in progress (2023).
\newblock \href {http://arxiv.org/abs/arXiv:2308.13916} {\path{arXiv:arXiv:2308.13916}}.

\bibitem{Hsieh2023}
C.-Y. Hsieh, C.-L. Li, C.-K. Yeh, et~al., Distilling step by step! outperforming larger language models with less training data and smaller model sizes, Findings of the Association for Computational Linguistics: ACL 2023Accepted to Findings of ACL 2023 (2023).

\bibitem{Liu2024b}
A.~Liu, B.~Feng, et~al., Deepseek-v2: A strong, economical, and efficient mixture-of-experts language model, arXiv preprint (2024).
\newblock \href {http://arxiv.org/abs/arXiv:2405.04434} {\path{arXiv:arXiv:2405.04434}}.

\bibitem{Yang2024BB}
A.~Yang, B.~Zhang, B.~Hui, et~al., Qwen2.5-math technical report: Toward mathematical expert model via self-improvement, arXiv preprint (2024).
\newblock \href {http://arxiv.org/abs/arXiv:2409.12122} {\path{arXiv:arXiv:2409.12122}}.

\bibitem{Deroy2024}
A.~Deroy, S.~Maity, Code generation and algorithmic problem solving using llama 3.1 405b, arXiv preprintUnder Review (2024).
\newblock \href {http://arxiv.org/abs/arXiv:2409.19027} {\path{arXiv:arXiv:2409.19027}}.

\bibitem{Papineni2002}
K.~Papineni, S.~Roukos, T.~Ward, W.-J. Zhu, Bleu: A method for automatic evaluation of machine translation, in: Proceedings of the 40th Annual Meeting of the Association for Computational Linguistics, 2002, pp. 311--318.

\bibitem{Lin2003}
C.-Y. Lin, E.~Hovy, Automatic evaluation of summaries using n-gram co-occurrence statistics, in: Proceedings of the 41st Annual Meeting of the Association for Computational Linguistics (ACL), Sapporo, Japan, 2003, pp. 707--712.

\bibitem{smith2022humanevaluationconversationsopen}
E.~M. Smith, O.~Hsu, R.~Qian, S.~Roller, Y.-L. Boureau, J.~Weston, \href{https://arxiv.org/abs/2201.04723}{Human evaluation of conversations is an open problem: comparing the sensitivity of various methods for evaluating dialogue agents} (2022).
\newblock \href {http://arxiv.org/abs/2201.04723} {\path{arXiv:2201.04723}}.
\newline\urlprefix\url{https://arxiv.org/abs/2201.04723}

\bibitem{zheng2023judgingllmasajudgemtbenchchatbot}
L.~Zheng, W.-L. Chiang, Y.~Sheng, S.~Zhuang, Z.~Wu, Y.~Zhuang, Z.~Lin, Z.~Li, D.~Li, E.~P. Xing, H.~Zhang, J.~E. Gonzalez, I.~Stoica, \href{https://arxiv.org/abs/2306.05685}{Judging llm-as-a-judge with mt-bench and chatbot arena} (2023).
\newblock \href {http://arxiv.org/abs/2306.05685} {\path{arXiv:2306.05685}}.
\newline\urlprefix\url{https://arxiv.org/abs/2306.05685}

\bibitem{Mese2025}
M.~I, K.~B, Chatgpt as an effective tool for quality evaluation of radiomics research, European Radiology 35~(4) (2025) 2030--2042.
\newblock \href {https://doi.org/10.1007/s00330-024-11122-7} {\path{doi:10.1007/s00330-024-11122-7}}.

\bibitem{Min2023}
S.~Min, K.~Krishna, X.~Lyu, M.~Lewis, W.~tau Yih, P.~Koh, M.~Iyyer, L.~Zettlemoyer, H.~Hajishirzi, Factscore: Fine-grained atomic evaluation of factual precision in long form text generation, in: Proceedings of the 2023 Conference on Empirical Methods in Natural Language Processing, 2023, pp. 12076--12100.

\end{thebibliography}
\end{document}